\documentclass[numbers]{article} 
\usepackage{iclr2025_conference,times}

\usepackage{amsmath,amsfonts,bm}

\def\eqref#1{equation~\ref{#1}}

\def\1{\bm{1}}

\DeclareMathAlphabet{\mathsfit}{\encodingdefault}{\sfdefault}{m}{sl}
\SetMathAlphabet{\mathsfit}{bold}{\encodingdefault}{\sfdefault}{bx}{n}

\usepackage{hyperref}
\usepackage{url}

\usepackage{amsfonts}
\usepackage{amsmath}
\usepackage{booktabs}
\usepackage{multirow}
\usepackage{graphicx}
\usepackage{enumitem}
\usepackage{textcomp}
\usepackage{booktabs}
\usepackage{subfigure}
\usepackage{array}
\usepackage{bm}
\usepackage{tabularx}
\usepackage{tabulary}
\usepackage{array}
\usepackage{color,ulem}
\usepackage{xcolor}
\usepackage{caption}
\usepackage{makecell} 
\usepackage{xcolor} 
\usepackage{fontawesome5} 
\usepackage{hyperref}

\usepackage{newtxtext}  
\title{Klear-Reasoner: Advancing Reasoning Capability via Gradient-Preserving Clipping Policy Optimization}
\author{
Zhenpeng Su\footnotemark[1] \quad Leiyu Pan  \quad Xue Bai \quad Dening Liu \quad Guanting Dong \quad  \textbf{Jiaming Huang} \\ \textbf{Minxuan Lv} \quad \textbf{Wenping Hu} \quad \textbf{Fuzheng Zhang} \quad \textbf{Kun Gai} \quad \textbf{Guorui Zhou} \footnotemark[2] \\  \\
  \textbf{Klear Team, Kuaishou Technology}  \\
}

\iclrfinalcopy 
\begin{document}

\maketitle
\renewcommand{\thefootnote}{\fnsymbol{footnote}} 
\footnotetext[1]{Project Leader.} \footnotetext[2]{Corresponding authors. }
\renewcommand{\thefootnote}{\arabic{footnote}}

\vspace{-0.7cm}
\begin{abstract}
\vspace{-0.3cm}
%
We present Klear-Reasoner, a model with long reasoning capabilities that demonstrates careful deliberation during problem solving, achieving outstanding performance across multiple benchmarks. Although there are already many excellent works related to inference models in the current community, there are still many problems with reproducing high-performance inference models due to incomplete disclosure of training details.
This report provides an in-depth analysis of the reasoning model, covering the entire post-training workflow from data preparation and  long Chain-of-Thought supervised fine-tuning (long CoT SFT) to reinforcement learning (RL), along with detailed ablation studies for each experimental component.  
For SFT data, our experiments show that a small number of high-quality data sources are more effective than a large number of diverse data sources, and that difficult samples can achieve better results without accuracy filtering.
In addition, we investigate two key issues with current clipping mechanisms in RL: Clipping suppresses critical exploration signals and ignores suboptimal trajectories.
To address these challenges, we propose \textbf{G}radient-\textbf{P}reserving clipping \textbf{P}olicy \textbf{O}ptimization (GPPO) that gently backpropagates gradients from clipped tokens. 
GPPO not only enhances the model's exploration capacity but also improves its efficiency in learning from negative samples. 
Klear-Reasoner exhibits exceptional reasoning abilities in mathematics and programming, \textbf{scoring 90.5\% on AIME 2024, 83.2\% on AIME 2025, 66.0\% on LiveCodeBench V5 and 58.1\% on LiveCodeBench V6}. 


\end{abstract}

\begin{center}
    \scalebox{0.89}{
        \includegraphics[width=0.88\textwidth]{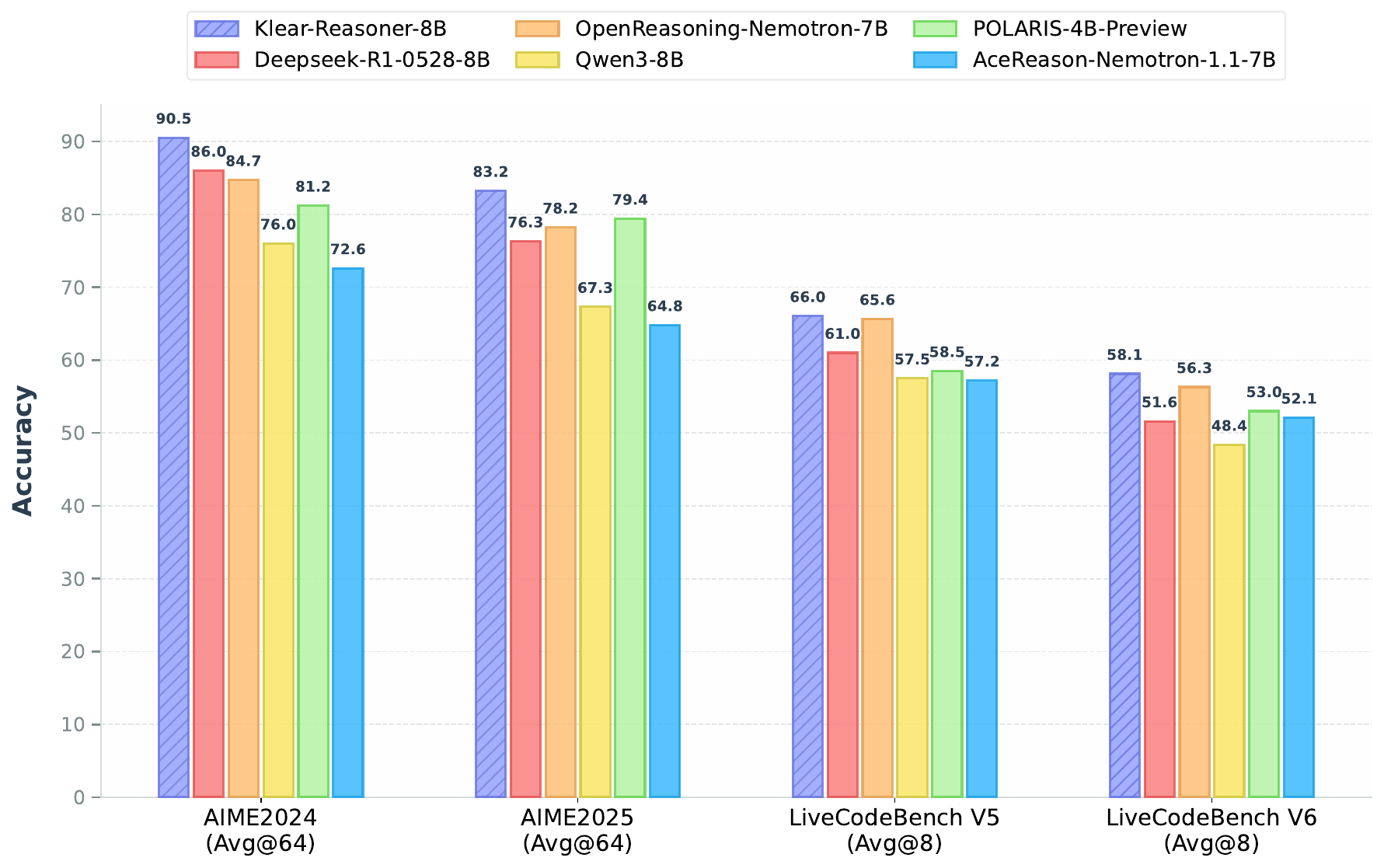}
    }
\end{center}
\captionof{figure}{Benchmark accuracy of Klear-Reasoner-8B on AIME 2024/2025 (avg@64), LiveCodeBench V5 (2024/08/01-2025/02/01, avg@8), and v6 (2025/02/01-2025/05/01, avg@8).}
\label{fig:main_results}

\section{Introduction}

The demonstrated success of OpenAI's O1 series models~\citep{DBLP:journals/corr/abs-2412-16720} and DeepSeek-R1~\citep{DBLP:journals/corr/abs-2501-12948} highlights the significant potential of large-scale reinforcement learning for complex reasoning tasks. However, due to the incomplete disclosure of training details, many deep-seated issues remain regarding the reproduction of high-performance reasoning models. Although the community has produced numerous excellent reproductions of small models with 7–8B parameters, such as DeepScaleR~\citep{deepScaler2025,deepcoder2025}, Light-R1~\citep{DBLP:journals/corr/abs-2503-10460}, Ring-7B~\citep{DBLP:journals/corr/abs-2506-14731}, AReal-boba-RL-7B~\citep{fu2025areal} and AceReason-Nemotron-1.1-7B~\citep{DBLP:journals/corr/abs-2506-13284}, their performance in mathematical and coding reasoning tasks still lags behind DeepSeek-R1-0528-Distill-8B~\citep{DBLP:journals/corr/abs-2501-12948} and Qwen3-8B~\citep{DBLP:journals/corr/abs-2505-09388}.

In this report, we conduct a detailed analytical study of Qwen3-8B-Base~\citep{DBLP:journals/corr/abs-2505-09388} to advance its reasoning capabilities to state-of-the-art levels for models of comparable scale.
We discover that for long Chain-of-Thought supervised fine-tuning (long CoT SFT), a compact set of high-quality data sources proves significantly more effective than larger, more diverse datasets, as high-quality examples ensure consistent learning of accurate reasoning patterns. Then, we find that simple SFT samples that are not filtered for correctness can easily interfere with the model, but difficult samples do not appear to harm the model's performance even if they are not filtered. In fact, these difficult errors may even promote the model's exploration and are beneficial to its performance.

For reinforcement learning, clipping importance sampling is a commonly used technique. The clipping mechanism primarily controls the magnitude of policy model updates to ensure training stability. Through in-depth analysis, we identify two issues with the current clipping mechanism~\citep{DBLP:journals/corr/SchulmanWDRK17}:
\begin{itemize}[leftmargin=*]
\item \textbf{High-entropy token clipping.} Among the tokens clipped beyond the upper threshold \(1+\epsilon\) of importance sampling, there exist high-entropy tokens that often correspond to valuable exploratory behaviors at critical decision points. Directly clipping these tokens may lead to premature termination of exploration, adversely affecting the model's post-convergence performance. Although DAPO~\citep{DBLP:journals/corr/abs-2503-14476} proposes Clip-Higher to mitigate this issue by adjusting the upper importance sampling threshold to $1+\epsilon_h$, high-entropy tokens exceeding this threshold still face the same problem. 
\item \textbf{Delayed convergence of negative samples.} When the importance sampling ratio of suboptimal trajectories falls below \(1-\epsilon\), their gradients are forcibly truncated, preventing the model from updating based on these signals, which in turn slows down convergence. 
\end{itemize}
To address the aforementioned two issues, we propose a \textbf{G}radient-\textbf{P}reserving clipping \textbf{P}olicy \textbf{O}ptimization (GPPO) that does not discard the gradients of any tokens. Even for truncated tokens, they are still included in the computational graph of backpropagation and participate in gradient computation. The gradients propagated back from these tokens by GPPO can be proven to be bounded and mild. This mechanism strikes a balance between maintaining training stability and preserving valuable gradient information. Our experimental results demonstrate that, compared to using Clip-Higher, GPPO achieves superior and more stable performance.

By integrating long CoT SFT with GPPO RL for math and coding tasks, we derive Klear-Reasoner-8B, a model that outperforms Qwen3-8B\footnote{https://huggingface.co/Qwen/Qwen3-8B} and DeepSeek-R1-0528-Distill-8B\footnote{https://huggingface.co/deepseek-ai/DeepSeek-R1-0528-Qwen3-8B} across key reasoning benchmarks, as shown in Figure \ref{fig:main_results}. Klear-Reasoner-8B achieves 90.5\% on AIME2024, 83.2\% on AIME2025, 66.0\% on LiveCodeBench V5, and 55.4\% on LiveCodeBench V6. Comprehensive ablations validate the efficacy of each component in our training pipeline.

\section{Preliminary}

Before introducing GPPO, we first provide a brief overview of  classical RL objectives and key features: 
\subsection{Proximal Policy Optimization (PPO)}
\label{Proximal Policy Optimization (PPO)}




PPO ~\citep{DBLP:journals/corr/SchulmanWDRK17} is a policy gradient method that stabilizes training via a clipped surrogate objective, constraining policy updates to avoid excessive deviations. Its core objective combines reward maximization with policy constraints through:

\begin{equation}
\begin{aligned}
\label{ppo_loss}
\mathcal{L}^{\text{PPO}}(\theta) = \mathbb{E}_{x\sim\mathcal{D}} \left[ \sum_{t=1}^{T} \min\left( r_t(\theta)\hat{A}_t, \text{clip}(r_t(\theta), 1-\epsilon, 1+\epsilon)\hat{A}_t \right) \right]
\end{aligned}
\end{equation}

Here, \( r_t(\theta) = \frac{\pi_\theta(a_t|s_t)}{\pi_{\text{old}}(a_t|s_t)} \) denotes the importance ratio, quantifying the policy change probability for action \(a_t\) in state \(s_t\) between the current policy \(\pi_\theta\) and the previous policy \(\pi_{\text{old}}\). The Generalized Advantage Estimate (GAE) \(\hat{A}_t\)~\citep{DBLP:journals/corr/SchulmanMLJA15} reduces variance in advantage estimation through discounted temporal differences:

\begin{equation}
\begin{aligned}
\hat{A}_t = \sum_{l=0}^{T-t} (\gamma\lambda)^l \delta_{t+l}, \quad \delta_t = r_t + \gamma V(s_{t+1}) - V(s_t)
\end{aligned}
\end{equation}

where \(\gamma\) is the discount factor controlling future reward weighting, \(\lambda\) is the GAE parameter balancing bias-variance trade-offs, \(V\) represents the value function, and \(r_t\) is the immediate reward. The clipping parameter \(\epsilon\), typically 0.2, enforces a hard constraint on policy updates by limiting \(r_t(\theta)\) to \([1-\epsilon, 1+\epsilon]\). The \(\min()\) operator implements a \textbf{pessimistic bound}: when advantages \(\hat{A}_t\) are positive, it clips overly optimistic updates; for negative advantages, it does not suppress the update magnitude. This mechanism ensures stable policy improvement within a trust region.

\subsection{Group Relative Policy Optimization (GRPO)}





GRPO~\citep{DBLP:journals/corr/abs-2402-03300} extends PPO by replacing value-based advantage estimation with group-normalized rewards, eliminating the need for a separate value network. For each prompt \(x\) and its group of \(M\) responses \(\{y^{(j)}\}_{j=1}^M\), the objective constrains policy updates through:

\begin{equation}
\begin{aligned}
\label{grpo_loss}
\mathcal{L}^{\text{GRPO}}(\theta) = \mathbb{E}_{x\sim\mathcal{D}} \left[ \frac{1}{M} \sum_{j=1}^{M} \frac{1}{|y^{(j)}|} \sum_{t=1}^{T_j} \min\left( r_t^{(j)}(\theta)\tilde{A}^{(j)}, \text{clip}\left(r_t^{(j)}(\theta), 1-\epsilon, 1+\epsilon\right)\tilde{A}^{(j)} \right) \right].
\end{aligned}
\end{equation}

Here, \( r_t^{(j)}(\theta) = \frac{\pi_\theta(a_t^{(j)}|s_t^{(j)})}{\pi_{\text{old}}(a_t^{(j)}|s_t^{(j)})} \) remains the importance ratio per token \(t\) in response \(j\), while \(\tilde{A}^{(j)}\) represents the group-relative advantage computed across all responses:  
\begin{equation}
\begin{aligned}
\tilde{A}^{(j)} = \frac{R^{(j)} - \mu_R}{\sigma_R}, \quad \mu_R = \frac{1}{M}\sum_{j=1}^M R^{(j)}, \quad \sigma_R = \sqrt{\frac{1}{M}\sum_{j=1}^M (R^{(j)}-\mu_R)^2}
\end{aligned}
\end{equation}
where \(R^{(j)}\) is the cumulative reward for response \(j\), \(\mu_R\) is the group mean reward, and \(\sigma_R\) is the reward standard deviation within the response group. This normalization eliminates advantage bias across prompts. The length normalization factor \(\frac{1}{|y^{(j)}|}\), with \(|y^{(j)}|\) denoting response length, ensures equitable contribution for variable-length trajectories.

\subsection{Raise the Ceiling: Clip-Higher}
\label{Raise the Ceiling: Clip-Higher}



Standard PPO/GRPO employ symmetric clipping bounds $1 \pm \epsilon$, which can inadvertently suppress exploratory actions by equally constraining both advantageous and disadvantageous policy updates. To address this, DAPO~\citep{DBLP:journals/corr/abs-2503-14476} introduces asymmetric clipping, defined as:  

\begin{equation}
\begin{aligned}
\text{clip}_{\text{asym}}(r, \epsilon_l, \epsilon_h) = \begin{cases} 
1-\epsilon_l & \text{if } r < 1-\epsilon_l \\
1+\epsilon_h & \text{if } r > 1+\epsilon_h \\
r & \text{otherwise}
\end{cases}
\end{aligned}
\end{equation}

The modified objective function leverages these asymmetric bounds to balance exploration and conservatism:  

\begin{equation}
\begin{aligned}
\label{ppo_loss}
\mathcal{L}^{\text{clip-high}}(\theta) = \mathbb{E}_{x\sim\mathcal{D}} \left[ \frac{1}{M} \sum_{j=1}^{M} \frac{1}{|y^{(j)}|} \sum_{t=1}^{T} \min\left( r_t(\theta)\hat{A}_t, \text{clip}(r_t(\theta), 1-\epsilon_l, 1+\epsilon_h)\hat{A}_t \right) \right]
\end{aligned}
\end{equation}

where typically \(\epsilon_h > \epsilon_l\) (e.g., \(\epsilon_h=0.28\), \(\epsilon_l=0.2\)) to allow more aggressive exploration while maintaining conservative updates for unfavorable actions.

\subsection{Token-Level Policy Gradient Loss}
\label{Token-Level Policy Gradient Loss}

Recent work~\citep{DBLP:journals/corr/abs-2503-14476} has demonstrated that applying policy gradient methods at the token level, rather than the sample level, can yield significant improvements in training stability and final model performance. The token-level variant of the GRPO objective equipped with clip-higher can be expressed as:

\begin{equation}
\begin{aligned}
\label{grpo_loss_obj}
\mathcal{L}^{\text{GRPO}}(\theta) = \mathbb{E}_{x\sim\mathcal{D}} \left[ \frac{1}{\sum_{j=1}^M T_j} \sum_{j=1}^M \sum_{t=1}^{T_j} \min\left( r_t^{(j)}(\theta)\tilde{A}^{(j)}, \text{clip}\left(r_t^{(j)}(\theta), 1-\epsilon_l, 1+\epsilon_h\right)\tilde{A}^{(j)} \right) \right]
\end{aligned}
\end{equation}

where \( \sum_{j=1}^M T_j \) serves as the normalization factor across all tokens. Token-level policy loss ensures that each token contributes equally to the gradient updates rather than weighting entire samples uniformly, which enables more balanced optimization across variable-length reasoning chains and leads to significantly more stable policy updates, particularly for long trajectories where traditional sample-level averaging might dilute critical learning signals. The token-level granularity preserves fine-grained training signals throughout extended reasoning processes while maintaining consistent gradient magnitudes regardless of sequence length.

\section{Method}
In this section, we present the details of the supervised fine-tuning and reinforcement learning processes used to train Klear-Reasoner. 

\subsection{long Chain-of-Thought Supervised Fine-tuning}


For mathematical and coding tasks, we adopt a quality-centric data construction strategy inspired by OpenThoughts~\citep{DBLP:journals/corr/abs-2506-04178}, prioritizing high-quality data over superficial diversity. This approach is reflected in four key design principles: 
\begin{itemize}[leftmargin=*]
\item Our prompts are exclusively curated from high-quality sources: mathematics prompts primarily draw from OpenThoughts~\citep{DBLP:journals/corr/abs-2506-04178}, NuminaMath~\citep{numina_math_datasets}, and AceReason-Nemotron 1.1~\citep{DBLP:journals/corr/abs-2506-13284}, while coding prompts are gathered from OpenThoughts, OpenCodeReasoning~\citep{DBLP:journals/corr/abs-2504-01943}, AceReason-Nemotron 1.1, TACO~\citep{li2023taco}, Apps~\citep{hendrycksapps2021}, and Codeforces~\citep{penedo2025codeforces}.

\item To ensure data uniqueness and prevent contamination, strict deduplication protocols are implemented, including exact match removal of queries and filtering prompts with 9-gram overlap against test samples following ~\citep{DBLP:journals/corr/abs-2505-16400, DBLP:journals/corr/abs-2506-13284}.

\item The teacher model employed for response generation is Deepseek-R1-0528\footnote{https://huggingface.co/deepseek-ai/DeepSeek-R1-0528}, which produces up to 16 responses per prompt through sampling.

\item After evaluating sample difficulty using Qwen3-8B~\citep{DBLP:journals/corr/abs-2505-09388}, we retain all responses since most samples qualify as difficult, consistent with Section \ref{section_5_1_1} insights.
\end{itemize}
Through this process, we construct a high-quality reasoning dataset containing 1.5 million math and coding samples.

To effectively leverage these high-quality reasoning data, we fine-tune the model using the standard SFT objective:
\begin{equation}
\begin{aligned}
\mathcal{L}^{\text{LM}}(\theta) = -\frac{1}{N} \sum_{j=1}^{N} \sum_{t=1}^{T_j} \pi_\theta(a_t|s_t)
\end{aligned}
\end{equation}

where \(\theta\) denotes model parameters, \(N\) is the total training sequences, \(T_j\) is the token count of sequence \(j\), \(a_t\) is the target token at position \(t\), and \(s_t\) is the contextual history. The term \(\pi_\theta(a_t | s_t)\) represents the model's probability of generating the true token \(a_t\) given context \(s_t\). By minimizing the negative average probability across all tokens and sequences, this objective aligns the model's output distribution with high-quality teacher demonstrations.

\subsection{Reinforcement Learning}
\subsubsection{Data curation}
We aim to filter high-quality reasoning data for RL. To this end, we have developed a data collection and validation pipeline that integrates prompts from various sources, including Skywork-OR1~\citep{DBLP:journals/corr/abs-2505-22312}, Acereason~\citep{DBLP:journals/corr/abs-2505-16400}, NuminaMath~\citep{numina_math_datasets}, and DeepScaleR~\citep{deepScaler2025,deepcoder2025}. A 9-gram filtering mechanism is employed to prevent overlap with common benchmark datasets~\citep{DBLP:journals/corr/abs-2505-16400, DBLP:journals/corr/abs-2506-13284}, and exact deduplication is applied to ensure the uniqueness of each sample.
For code problems, we apply multiple filtering strategies to mitigate noisy or low-quality samples that could harm RL training. We first remove examples with fewer than 16 test cases, as these are more susceptible to false positives. For each retained prompt, we generate 16 completions using DeepSeek-R1-0120\footnote{https://huggingface.co/deepseek-ai/DeepSeek-R1}. Only those prompts for which DeepSeek-R1-0120 achieves a $pass@16$ greater than 0.5 are kept, reducing noise introduced by faulty or insufficient test cases.
For math problems, we focus on ensuring correctness and clarity, particularly because the raw data, often collected from the web and processed via OCR or parsers, can contain substantial noise (e.g., incorrect questions or answers). To address this, we generate 16 responses per prompt using DeepSeek-R1-0120 and retain only those for which a majority of completions pass a rule-based validator.
Finally, after these rigorous filtering steps, we construct a high-quality RL dataset consisting of 88K math samples and 18K code samples.


\subsubsection{Gradient-Preserving Clipping Policy Optimization}\label{sec:Gradient-Preserving}

In the classical PPO algorithm~\citep{DBLP:conf/icml/SchulmanLAJM15,DBLP:journals/corr/SchulmanWDRK17} without clipping, we can derive the gradient of the loss function as shown in Eq.\ref{ppo_gradient}. Here, $f(a,s)$ denotes the logits output by the policy network. \( b \) iterates over all possible actions in the action space \( \mathcal{A} \) at state \( s_t \).

\begin{equation}
\label{ppo_gradient}
\begin{aligned}
\nabla_\theta \mathcal{L}^{\text{PPO}} &= \mathbb{E}_t \left[ r_t(\theta) \cdot \phi_\theta(a_t, s_t) \cdot \hat{A}_t \right], \\
\text{where } \phi_\theta(a_t, s_t) &= \frac{\partial f_\theta(a_t, s_t)}{\partial \theta} - \sum_{b \in \mathcal{A}} \pi_\theta(b|s_t) \cdot \frac{\partial f_\theta(b, s_t)}{\partial \theta}
\end{aligned}
\end{equation}
However, the absence of clipping allows the importance sampling ratio \( r_t(\theta) \) to range from \( (0, +\infty) \). While this broad range enhances exploration by enabling significant policy updates in promising directions, such overly aggressive exploration may lead to training instability due to excessively large gradient updates. This can result in oscillating or divergent policy updates, ultimately hindering convergence to the optimal policy.  

To mitigate the aforementioned issues, a common approach is to apply clipping to the importance sampling ratio, constraining its upper and lower bounds, as shown in Eq.\ref{ppo_loss} and Eq.\ref{grpo_loss}. Similarly, we can derive the gradient expression after applying the clipping operation, as presented in Eq.\ref{ppo_clip_gradient}.

\begin{equation}
    \begin{aligned}
    \label{ppo_clip_gradient}
        \nabla_\theta \mathcal{L}^{\text{PPO-clip}} &= \mathbb{E}_t \left[ \mathbb{I}_t \cdot r_t(\theta) \cdot \phi_\theta(a_t, s_t) \cdot \hat{A}_t \right] \\[0.8em]
        \text{where} \quad \mathbb{I}_{t} &= \begin{cases}
            1, & \text{if } r_{t}(\theta) \cdot \hat{A}_t \le \text{clip}(r_t(\theta), 1 - \epsilon, 1 + \epsilon) \cdot \hat{A}_t \\[0.3em]
            0, & \text{otherwise}
        \end{cases}
    \end{aligned}
\end{equation}

From the above equation, we can infer the following: when $\hat{A}_t > 0$, gradients are only propagated if the importance sampling ratio $r_t(\theta)$ falls within the range $(0, 1 + \epsilon)$; when $\hat{A}_t < 0$, gradients are only present if $r_t(\theta)$ lies within $(1 - \epsilon, +\infty)$. Compared to the unclipped case, this clipping mechanism effectively restricts the range of gradient updates, capping the positive advantage updates to prevent over-encouragement, and flooring the negative advantage updates to avoid excessive punishment. 
Although introducing clip operations stabilized training to a certain extent, completely discarding the clipped token gradients also led to two key issues.


\begin{itemize}[leftmargin=*]
    \item \textbf{High-entropy token clipping.} A key side effect of clipping is the indiscriminate suppression of gradients for all tokens whose importance sampling ratios exceed predefined thresholds $1+\epsilon$, regardless of whether those tokens may contain valuable exploratory behaviors. In our preliminary experiments, we observed a phenomenon similar to that reported in ~\citep{DBLP:journals/corr/abs-2503-14476,DBLP:journals/corr/abs-2506-13585}: high-entropy tokens associated with reflective behaviors often lie at crucial decision branches. However, due to their elevated $r_t^{(j)}(\theta)$ values, these tokens are typically clipped out after the first on-policy update, severely hampering the model’s capacity to explore.
    \item \textbf{Delayed convergence of negative samples.} When the importance sampling ratio of suboptimal trajectories falls below $1-\epsilon$, their gradients are forcibly clipped, preventing the model from updating based on these signals. As a result, the model must repeatedly sample similar suboptimal trajectories before it can correct its behavior, leading to slower convergence on negative examples and hindering timely policy adjustment.
\end{itemize}


Existing methods incorporating gradient signals from beyond clip boundaries, such as DAPO’s clip-higher operation that elevates the clip upper bound to preserve high-entropy tokens, fail to resolve the underlying issue and still inherently truncate gradient signals from certain samples.
Fundamentally, the issue described above also stems from side effects induced by the clip operation. \textbf{Therefore, the core motivation of our method is to propose a balanced approach that bridges unclipped and clipped methodologies. This approach aims to incorporate gradient signals from samples beyond clip boundaries while constraining these out-of-bound gradient signals within a defined range.} Such constraint prevents them from exerting excessive influence on gradient updates, thereby safeguarding training stability. 


To address the above challenges, we propose a \textbf{G}radient-\textbf{P}reserving clipping \textbf{P}olicy \textbf{O}ptimization (GPPO).  Unlike traditional techniques that completely discard gradients outside the clipping range, our method introduces an innovative truncation mechanism that incorporates previously clipped tokens into the computational graph. This design ensures that valuable learning signals are retained while maintaining stable policy updates. Specifically, taking GRPO loss as an example, we modify Eq.\ref{grpo_loss_obj} as follows:

\begin{equation}
\begin{aligned}
\label{grpo_loss_soft}
\mathcal{L}^{\text{GPPO}}(\theta) &= \mathbb{E}_{x\sim\mathcal{D}} \left[ \frac{1}{\sum_{j=1}^M T_j} \sum_{j=1}^M \sum_{t=1}^{T_j} \min\left( \delta\tilde{A}^{(j)}, \text{clip}\left(\delta, \frac{1-\epsilon_l}{\operatorname{sg}(\delta)} \delta, \frac{1+\epsilon_h}{\operatorname{sg}(\delta)} \delta\right)\tilde{A}^{(j)} \right) \right]
\end{aligned}
\end{equation}
Here, $\delta = r_t^{(j)}(\theta)$ is the importance sampling ratio, and $\operatorname{sg}(\cdot)$ denotes the stop-gradient operation. Notably, \textbf{the term $\frac{\delta}{\operatorname{sg}(\delta)}$ is always numerically equal to 1, so the forward computation remains unchanged}. Since we decouple gradient propagation from the clipping constraint, the backward computation differs from the standard clipping approach, which directly discards gradients of samples falling outside the clipping bounds. The gradient expression of our proposed method is presented in Eq.\ref{grpo_gradient_soft}.



\begin{equation}
\begin{aligned}
\label{grpo_gradient_soft}
\nabla_\theta \mathcal{L}^{\text{GPPO}}(\theta) &= \mathbb{E}_{x \sim \mathcal{D}} \left[
\frac{1}{\sum_{j=1}^M T_j} \sum_{j=1}^M \sum_{t=1}^{T_j}
\mathcal{F}_{j,t}(\theta) \cdot \phi_\theta(a_{j,t}, s_{j,t}) \cdot \tilde{A}^{(j)}
\right], \\[1.2em]
\text{where} \quad 
\mathcal{F}_{j,t}(\theta) &= 
\begin{cases}
1-\epsilon_l & \text{if } \delta < 1-\epsilon_l \text{ and } \tilde{A}^{(j)} < 0, \\[0.5em]
1+\epsilon_h & \text{if } \delta > 1+\epsilon_h \text{ and } \tilde{A}^{(j)} > 0, \\[0.5em]
\delta & \text{otherwise (i.e., } \delta \tilde{A}^{(j)} \le \text{clip}(\delta, 1-\epsilon_l, 1+\epsilon_h) \cdot \tilde{A}^{(j)}).
\end{cases}
\end{aligned}
\end{equation}

By comparing Eq.\ref{ppo_clip_gradient} and Eq.\ref{grpo_gradient_soft}, we observe that our method avoids cases where the gradient becomes zero, meaning it fully leverages the gradient signals from all samples. Specifically, when the positive samples have a large importance sampling ratio $\delta > 1 + \epsilon_h$, it may correspond to high-entropy tokens. Our method preserves the gradients of these samples, thereby \textbf{enhancing the model’s exploratory capability}. Conversely, when the negative samples have $\delta < 1 - \epsilon_l$, the standard clipping mechanism discards its gradient, \textbf{leading to delayed convergence}. By retaining the gradient in these cases, our approach enables faster learning from negative samples, thus equipping the model with a stronger ability to perform rapid trial and error.

It is particularly noteworthy that although our method incorporates the gradient contributions from samples outside the clipping range, it still maintains stable policy updates. This is because for tokens clipped under the conditions $\delta < 1-\epsilon_l \text{ and } \tilde{A}^{(j)} < 0$, their backpropagated gradients are constrained to $1-\epsilon_l$, while for those clipped when $\delta > 1+\epsilon_h \text{ and } \tilde{A}^{(j)} > 0$, their gradients are constrained to $1+\epsilon_h$. 
In other words, for clipped tokens, we adjust their gradients in a controlled manner to ensure training stability, avoiding drastic gradient updates as described in Eq.\ref{ppo_gradient}. 
In summary, our approach not only accelerates the model's convergence when learning from negative samples but also facilitates exploration through high-entropy tokens, while maintaining training stability via moderate gradient backpropagation. A more general form of GPPO is shown in Appendix \ref{sec:The General Form of GPPO}.

Additionally, concurrent work by MiniMax-M1~\citep{DBLP:journals/corr/abs-2506-13585} explores a related direction through their proposed CISPO method. To enable gradient backpropagation on all tokens, CISPO adopts a vanilla REINFORCE objective with a corrected distribution for RL training. 
While Appendix~\ref{appendix:CISPO} reveals gradient similarities between CISPO and our method GPPO, their update strategies diverge fundamentally. GPPO adopts the pessimistic update principle established in Section~\ref{Proximal Policy Optimization (PPO)}: When the advantage \(\hat{A}_t\) is positive, GPPO suppresses overly optimistic updates, while CISPO does not impose any restrictions; Conversely, when the advantage is negative, GPPO imposes no restriction while CISPO does. Crucially, this distinction stems from differing design philosophies. GPPO derives a more principled clipping approach through gradient analysis, while CISPO leans more toward heuristic design.

\subsubsection{RL Training with SFT Loss}\label{sec:training_sft_loss}
Similar to~\citet{DBLP:journals/corr/abs-2504-05118}, we calculate the language model (LM) loss for the positive examples generated in each rollout and incorporate this loss into the RL training process. First, we argue that the SFT loss can improve the utilization efficiency of positive examples. Additionally, the SFT loss on positive examples serves as an anchor during training, constraining the policy model's output to maintain a reasonable distribution. We believe this helps mitigate reward hacking behavior and enhances training stability. Similarly, we employ a token-level loss, with the corresponding formula as follows:
\begin{equation}
\begin{aligned}
\mathcal{L}^{\text{LM}}(\theta) = -\frac{1}{ {\textstyle \sum_{j\in \Phi }}  T_j} \sum_{j\in \Phi } \sum_{t=1}^{T_j} \pi_\theta(a_t|s_t)
\end{aligned}
\end{equation}
Let $\Phi$ denote the index set of correct samples from examples in rollout. The final $\mathcal{L}^{\text{LM}}(\theta)$ is jointly trained with the GPPO loss, with the corresponding expression given by the following formula:

\begin{equation}
\begin{aligned}
\label{loss:eq}
\mathcal{L}^(\theta) = \mathcal{L}^{\text{GPPO}}(\theta)  + \alpha \mathcal{L}^{\text{LM}}(\theta)
\end{aligned}
\end{equation}
where $\alpha$ serves as a hyperparameter that balances the weighting between these two losses.

\subsubsection{Reward Design for Math and Code RL}
Our reinforcement learning framework employs distinct reward schemes for mathematical reasoning and code generation tasks. For mathematical tasks, we use a binary reward system where solutions receive either positive or negative rewards based on final correctness. Notably, we penalize responses that fail to encapsulate their reasoning process within designated \texttt{<think>...</think>} tags.

In terms of code tasks, RL often struggles with sparse rewards, particularly in code generation where models may produce largely correct solutions that fail only on corner cases. Traditional approaches label such samples as entirely negative, disregarding their partial correctness. We argue that these partially correct solutions contain valuable learning signals.
To overcome this challenge, we introduce a soft reward mechanism based on test case pass rates. For example, if generated code passes 4 out of 16 test cases, it receives a proportional reward of 0.25 (4/16). 
The soft $pass@k$ reward system provides granular feedback that mitigates sparse rewards, preserves learning signals from partially correct solutions, and encourages incremental quality improvement.

\section{Evaluation}
In this section, we present the final experimental results of Klear-Reasoner-8B. We first provide a detailed description of the training configuration, followed by an analysis of the training outcomes. Finally, we discuss the evaluation results.

\subsection{Training Details}
Here is a description of the training process for Klear-Reasoner-8B. Klear-Reasoner-8B is based on the Qwen3-8B-Base model, first fine-tuned with long CoT SFT and then trained using RL on math and coding tasks.  
For the training data, we collect math and coding problems from various sources and apply 9-gram filtering to remove data contaminated with evaluation sets, following \citep{DBLP:journals/corr/abs-2506-13284}. For the long CoT SFT data, we use DeepSeek-R1-0528 to generate responses for each question. Whether it is long CoT SFT or RL, we set the maximum training length to 32K. Following \citealp{DBLP:journals/corr/abs-2506-04178}, during long CoT SFT, we adopt a maximum learning rate of $8e^{-5}$ with a cosine decay strategy, decreasing to a minimum of $4e^{-5}$. We train for a total of 4 epochs to ensure model convergence.  
For RL, we employ a multi-stage training approach with a global batch size of 128 per step. Specifically, we adopt a two-stage training method, first training math RL and then training code RL.
We use an off-policy training strategy, with a constant learning rate of $1e^{-6}$ and a mini-batch size of 16 for math RL, and a constant learning rate of $5e^{-7}$ with a mini-batch size of 32 for coding RL. For rollout, we sample
8 responses for each prompt. It is worth mentioning that we use the pass rate of test cases as a reward for code RL. The RL training is conducted jointly with SFT loss, as described in Section \ref{sec:training_sft_loss}, where we set $\alpha$ to 0.1. Furthermore, we utilize the GPPO proposed in Section \ref{sec:Gradient-Preserving}. We apply only the token-level policy gradient loss described in Section \ref{Token-Level Policy Gradient Loss} and Clip-Higher with $\epsilon_h = 0.28$ as described in Section \ref{Raise the Ceiling: Clip-Higher}, following ~\citep{DBLP:journals/corr/abs-2505-22312}, without using KL loss during training.


\begin{table}[htbp]
\centering
\scalebox{0.95}{
\begin{tabular}{@{}lccccc@{}}
\toprule
\multirow{2}{*}{\textbf{Model}} & \multicolumn{5}{c}{\textbf{Benchmark}} \\
\cmidrule(lr){2-6}
& \textbf{\makecell{AIME2024  \\ \textcolor{gray}{avg@64}}} & \textbf{\makecell{AIME2025 \\\textcolor{gray}{avg@64}}} & \textbf{\makecell{HMMT2025  \\ \textcolor{gray}{avg@64}}} & \begin{tabular}{@{}c@{}}\textbf{\makecell{LCB V5 \\ \textcolor{gray}{avg@8}}} \end{tabular} & \begin{tabular}{@{}c@{}}\textbf{\makecell{LCB V6 \\ \textcolor{gray}{avg@8}}}\end{tabular} \\
\midrule
AReal-boba-RL-7B & 61.9 & 48.3 & 29.4 & 34.3 & 31.0$^\dagger$ \\
MiMo-7B-RL & 68.2 & 55.4 & 35.7 & 57.8 & 49.3 \\
Skywork-OR1-7B & 70.2 & 54.6 & 35.7 & 47.6 & 42.7 \\
AceReason-Nemotron-1.1-7B & 72.6 & 64.8 &42.9 & 57.2 & 52.1 \\
POLARIS-4B-Preview $^\spadesuit$ & 81.2 & \underline{79.4} & 58.7 & 58.5$^\dagger$ & 53.0$^\dagger$ \\
Qwen3-8B & 76.0 & 67.3 & 44.7$^\dagger$ & 57.5 & 48.4$^\dagger$ \\
Deepseek-R1-0528-Distill-8B  $^\clubsuit$ & \underline{86.0} & 76.3 & 61.5 & 61.0$^\dagger$ & 51.6$^\dagger$ \\
OpenReasoning-Nemotron-7B $^\clubsuit$ & 84.7 &78.2 & \underline{63.5} & \underline{65.6}$^\dagger$ & \underline{56.3}$^\dagger$
\\
\midrule
Klear-Reasoner-8B-SFT & 75.6 & 70.1 & 57.6 & 58.5 & 49.6 \\
Klear-Reasoner-8B & 83.2 & 75.6 & 60.3 & 61.6 & 53.1 \\
\quad \textit{w/ 64K Inference Budget} $^\clubsuit$ & \textbf{90.5} & \textbf{83.2} & \textbf{70.8} & \textbf{66.0} & \textbf{58.1} \\
\bottomrule
\end{tabular}
}
\caption{We report the average $pass@1$ results (avg@$n$), with all other evaluation metrics following the DeepSeek-R1 assessment framework (temperature=0.6, top\_p=0.95). LCB stands for LiveCodeBench. By default, we include official performance data provided by model developers when available. Otherwise, $\dagger$ indicates our evaluation results based on the officially recommended configurations. Models marked with $\spadesuit$ indicate a maximum inference length of 96K tokens, while those with $\clubsuit$ denote a 64K maximum inference length; all other models use a 32K inference length. The best score on a given dataset is marked in \textbf{bold}, and the second best is \underline{underlined}. }
\label{tab:transposed_benchmarks}
\vspace{0.2cm}
\end{table}

\subsection{Benchmarks and Baselines}
We evaluate the model on challenging benchmarks to assess its capabilities. For mathematical proficiency, we measure performance on AIME2024, AIME2025 as well as HMMT2025 Feb. For coding ability, we use LiveCodeBench V5 (covering August 2024 to February 2025) and LiveCodeBench V6 (spanning February 2025 to May 2025). We compare our model against several strong baselines, including OpenReasoning-Nemotron-7B~\citep{ahmad2025opencodereasoningiisimpletesttime}, Deepseek-R1-0528-Distill-8B~\citep{DBLP:journals/corr/abs-2501-12948}, Qwen3-8B~\citep{DBLP:journals/corr/abs-2505-09388}, POLAR-4B-Preview~\citep{Polaris2025}, AceReason-Nemotron-1.1-7B~\citep{DBLP:journals/corr/abs-2506-13284} and MiMo-7B-RL~\citep{coreteam2025mimounlockingreasoningpotential}, AReal-boba-RL-7B~\citep{fu2025areal} and Skywork-OR1~\citep{DBLP:journals/corr/abs-2505-22312}.


\subsection{Evaluation Results}
As shown in Table \ref{tab:transposed_benchmarks}, we present the evaluation results on math and code benchmarks. We have made the following results:
\begin{itemize}[leftmargin=*]
\item \textbf{SFT Model Performance.} Klear-Reasoner-8B-SFT model achieves performance comparable to Qwen3-8B. Notably, while Qwen3-8B is obtained through logits distillation from a larger model, Klear-Reasoner-8B-SFT is trained solely through data distillation (i.e., long CoT SFT), demonstrating the effectiveness of our long CoT SFT approach. 

\item \textbf{RL Model Performance.} Building upon SFT model, we further refine Klear-Reasoner-8B through RL fine-tuning for math and code tasks. With a 32K inference budget, Klear-Reasoner-8B already matches the performance of community SOTA models operating at 64K/96K inference budgets, achieving specific scores of 83.2\% on AIME2024, 75.6\% on AIME2025, 60.3\% on HMMT2025, 61.6\% on LiveCodeBench V5, and 53.1\% on LiveCodeBench V6.  

\item \textbf{More Reasoning Budget.} When we expand the inference budget to 64K and adopt the YaRN~\citep{DBLP:conf/iclr/PengQFS24} method with a scaling factor of 2.5, Klear-Reasoner-8B achieves optimal performance, scoring 90.5\% on AIME2024, 83.2\% on AIME2025, 70.8\% on HMMT2025, 66.0\% on LiveCodeBench V5, and 58.1\% on LiveCodeBench V6. It is worth noting that regardless of whether it is long CoT SFT or RL, our model is only trained on 32K length. 

\item \textbf{RL Fine-Tuning Surpasses Data-Heavy SFT.} OpenReasoning-Nemotron-7B is derived from Qwen2.5-7B using 5 million long CoT distillation samples from DeepSeek-R1-0528-Distill-8B. Although Klear-Reasoner-8B-SFT, trained with fewer data samples, underperforms OpenReasoning-Nemotron-7B~\citep{ahmad2025opencodereasoningiisimpletesttime}, the RL-trained Klear-Reasoner-8B with a 64K inference budget still significantly surpasses OpenReasoning-Nemotron-7B. This validates that RL fine-tuning can effectively compensate for the disadvantage of limited training data and further boost model performance. 
\end{itemize}

\begin{table}[!t]
\centering
\scalebox{1}{
\renewcommand{\arraystretch}{1.1}
\begin{tabular}{lcccccc}
\toprule
\textbf{Benchmark} & \multicolumn{2}{c}{\textbf{Easy}} & \multicolumn{2}{c}{\textbf{Hard}} & \multicolumn{2}{c}{\textbf{Overall}} \\
\cmidrule(lr){2-3} \cmidrule(lr){4-5} \cmidrule(lr){6-7}
 & Only True & Mixed & Only True & Mixed & Only True & Mixed \\
\midrule
AIME 2024       & \textbf{45.00}                 & 40.22                   & 45.63                & \textbf{47.29}                   & \textbf{44.58}                & 42.92                   \\
AIME 2025       & \textbf{31.25}                 & 30.63                   & 33.13                & \textbf{33.54}                   & 32.29                & \textbf{32.50}                   \\
LiveCodeBench V5  & \textbf{20.79}                 & 20.07                   & 25.81                & \textbf{26.88}                   & \textbf{24.73}                & 23.66                   \\
\bottomrule
\end{tabular}
}
\caption{Impact of data correctness on model performance across different benchmarks. Each group compares results on correct data (\textbf{Only True}) versus mixed data containing both correct and incorrect data (\textbf{Mixed}), evaluated over the \textit{easy}, \textit{hard} and \textit{overall} subsets. Experiments were conducted on balanced subsets (selected via K-Center Greedy) from the \texttt{OpenR1-Math-220k} dataset. \textbf{Bolded values} indicate the best performance within each group (\textit{easy}, \textit{hard} or \textit{overall}).}
\label{tab:data_correctness_impact}
\end{table}

\begin{table}[!t]
\centering
\scalebox{1}{
\renewcommand{\arraystretch}{1.1}
\begin{tabular}{lccc}
\toprule
\textbf{Benchmark} & \textbf{QwQ-32B} & \textbf{DeepSeek-R1-0120} & \textbf{DeepSeek-R1-0528} \\
\midrule
AIME 2024         & 62.71              & 67.71                     & \textbf{73.54}                          \\
AIME 2025         & 53.13              & 50.00                     & \textbf{64.58}                          \\
LiveCodeBench V5  & 44.44              & 50.90                     & \textbf{53.05}                          \\
\bottomrule
\end{tabular}
}
\caption{Impact of different teacher models on supervised fine-tuning performance. \textbf{Bolded values} indicate the best performance across teacher models. Note that results under DeepSeek-R1-0528 correspond to an early-stage SFT model.}
\label{tab:teacher_model_impact}
\end{table}

These results indicate that even when starting from a high-performance SFT model, a well-designed RL approach can further enhance model capabilities. 

\section{Analysis}
In this section, we present a detailed analytical study of long-chain-of-thought SFT and RL methods. The SFT model used in these experiments differs from the one employed in the previously introduced Klear-Reasoner-8B-SFT, and is instead based on an earlier version of our SFT model (as reported in Table~\ref{tab:teacher_model_impact}). Throughout all evaluations, we maintain strict control over individual variables to ensure valid and fair comparisons.

\subsection{SFT Analyses}
\subsubsection{Impact of data correctness}
\label{section_5_1_1}


Understanding the correctness of CoT samples is essential for effective model training. Our study leverages the \texttt{OpenR1-Math-220k} dataset\footnote{https://huggingface.co/datasets/open-r1/OpenR1-Math-220k}, which provides explicit correctness labels, and reveals that the impact of data correctness on performance is largely dependent on task difficulty, challenging the conventional belief that using only correct data is always optimal.

Initially, when comparing SFT performance on carefully curated, balanced subsets of the full dataset, we observe no statistically significant difference between training solely on correct CoT data and using a mixture of correct and incorrect responses. For example, as shown in the ``Overall" column of Table~\ref{tab:data_correctness_impact}, the performance gap is minimal: AIME 2025 benchmark yields 32.29\% for correct-only training versus 32.50\% for mixed data.

The critical insight emerges when we examine the data through the lens of difficulty. By stratifying the dataset into easy and hard subsets based on model-annotated difficulty labels, a striking divergence become apparent. \textbf{For easy tasks, training solely on correct data consistently yield superior results.} For instance, on the AIME 2024 benchmark, using only correct data resulted in a 4.78\% performance improvement compared to training with mixed responses. This suggests that when the underlying reasoning is relatively straightforward and aligns well with the model's prior capabilities, exposure to incorrect paths acts primarily as detrimental noise, interfering with the model's grasp of established, correct patterns.

\begin{table}[!t]
\centering
\scalebox{1}{
\renewcommand{\arraystretch}{1.1}
\begin{tabular}{lcccccc}
\toprule
\textbf{Benchmark}  & \textbf{Top1}  & \textbf{Top2}  & \textbf{Top4}  & \textbf{Top6}  & \textbf{Top8}  & \textbf{Top10} \\ \midrule
AIME 2024         & \textbf{40.83} & 37.71 & 37.29 & 33.96 & 29.38 & 30.42 \\ 
AIME 2025         & \textbf{36.04} & 31.86 & 29.79 & 29.17 & 28.96 & 30.63 \\ 
LiveCodeBench V5  & \textbf{24.37} & 20.79 & 19.80 & 13.35 & 19.89 & 14.70 \\ 
LiveCodeBench V6  & 23.47 & \textbf{23.85} & 20.61 & 14.50 & 19.66 & 14.31 \\
\bottomrule
\end{tabular}
}
\caption{Impact of combining top-K high-quality \textbf{math} data subsets on SFT performance. For each Top-K setting, we construct a training set by sampling 31,600 examples using the K-Center greedy algorithm. \textbf{Bolded values} indicate the best performance under each Top-K setting.}
\label{tab:data_quality_and_diversity_balance_math}
\end{table}

\begin{table}[!t]
\centering
\scalebox{1}{
\renewcommand{\arraystretch}{1.1}
\begin{tabular}{lcccccc}
\toprule
\textbf{Benchmark}  & \textbf{Top1}  & \textbf{Top2}  & \textbf{Top4}  & \textbf{Top6} \\ \midrule
AIME 2024         & \textbf{31.25} & 28.33 & 27.71 & 28.33  \\ 
AIME 2025         & \textbf{27.50} & 25.42 & 23.33 & 22.29  \\ 
LiveCodeBench V5  & 26.52 & \textbf{27.24} & 26.97 & 25.18  \\ 
LiveCodeBench V6  & 26.53 & \textbf{29.20} & 28.24 & 25.38  \\
\bottomrule
\end{tabular}
}
\caption{Impact of combining top-K high-quality \textbf{code} data subsets on SFT performance. For each Top-K setting, we construct a training set by sampling 31,600 examples using the K-Center greedy algorithm. \textbf{Bolded values} indicate the best performance under each Top-K setting.}
\label{tab:data_quality_and_diversity_balance_code}
\end{table}

\textbf{In contrast, for hard tasks, incorporating incorrect examples alongside correct ones surprisingly improved performance.} On the AIME 2024 Hard subset, training with mixed data achieved a 1.66\% higher accuracy than using only correct samples. This counterintuitive finding indicates that in scenarios characterized by high uncertainty and weaker initial learning signals, incorrect examples serve a valuable purpose. They provide necessary contrast, helping the model more effectively discriminate between valid and invalid reasoning paths. This mechanism may enhance the model's exploratory capacity within the solution space, preventing it from becoming trapped in suboptimal local reasoning patterns. This dynamic parallels the classic ``exploration vs. exploitation" trade-off observed in reinforcement learning.

This difficulty-dependent effect directly inform our final data strategy. An analysis of the full training dataset shows that a substantial majority (88\%) of samples are labeled as hard. Given this pronounced bias towards challenging examples and our empirical finding that mixed data benefits learning on such tasks, we deliberately choose not to filter CoT samples based on correctness during the training phase. The presence of incorrect reasoning, within a predominantly hard dataset, is leveraged as a beneficial learning signal rather than treated as noise to be eliminated.

\subsubsection{Ablation of teacher models}

In this section, we conduct an ablation study to evaluate how the choice of teacher model affects the effectiveness of supervised fine-tuning. Specifically, we use a fixed set of prompts from the \texttt{OpenThoughts3-1.2M} dataset\footnote{https://huggingface.co/datasets/open-thoughts/OpenThoughts3-1.2M} and generate responses using three different teacher models: QwQ-32B, DeepSeek-R1-0120, and DeepSeek-R1-0528.

From Table~\ref{tab:teacher_model_impact}, we observe a clear performance gap across the teacher variants. Distillation data generated by DeepSeek-R1-0528, the strongest teacher in our setup, leads to the highest student performance across benchmarks. In contrast, data from QwQ-32B yields the weakest results. These differences highlight the crucial role of teacher quality in distillation-based SFT.

Stronger teacher models tend to produce more accurate, coherent, and structured responses, which offer richer training signals. This in turn improves the student model's ability to generalize. Weaker teachers, on the other hand, may generate ambiguous or incomplete answers, introducing noise and degrading downstream performance. This confirms that teacher quality is a bottleneck in distillation, and selecting a capable teacher is essential for effective knowledge transfer.

\subsubsection{Data quality and diversity balance}



When selecting training data, a common trade-off arises between data quality and diversity. High-quality data often originates from a limited number of sources, which constrains its diversity. Conversely, incorporating data from more diverse sources can degrade overall data quality. In this section, we investigate how to balance quality and diversity in long-chain-of-thought datasets for mathematics and code reasoning tasks. We follow the data quality rankings provided by OpenThoughts ~\citep{DBLP:journals/corr/abs-2506-04178}. Experimental results for math and code tasks are presented in Table \ref{tab:data_quality_and_diversity_balance_math} and Table \ref{tab:data_quality_and_diversity_balance_code}, respectively.

From the results, we observe a clear trend: models fine-tuned on only the Top1 or Top2 high-quality sources achieve the best performance. As more data sources are added, performance consistently degrades. This pattern holds across both math and code domains, and aligns with findings reported in the OpenThoughts. We attribute this phenomenon to two primary factors:

\begin{itemize}[leftmargin=*]
    \item High-quality sources tend to encapsulate the most effective, internally consistent reasoning patterns necessary for solving complex tasks such as mathematics and code generation. With a fixed training budget, models benefit more from focusing on such core reasoning modes.
    \item The marginal utility of adding lower-quality sources is negative. These sources inevitably introduce noise, such as logical gaps, incorrect derivations, or inefficient problem-solving strategies, which competes for limited model capacity during training. Additionally, heterogeneous sources may exhibit conflicting reasoning styles, further disrupting learning and reducing training efficiency.
\end{itemize}

\textbf{In summary, to achieve strong performance in long CoT reasoning, prioritizing data quality and internal consistency is more effective than simply maximizing surface-level diversity.}
Based on the above findings, our long CoT SFT data is primarily sourced from a few high-quality datesets rather than broad and diverse datasets. Specifically, the mathematical data is mainly derived from OpenThoughts, while the coding data primarily comes from OpenCodeReasoning, TACO, Apps, and Codeforces.

\subsection{RL Analyses}

\begin{figure*}
    \centering
    \scalebox{0.95}{
    \subfigure[AIME2024]{
        \includegraphics[width=0.33\textwidth]{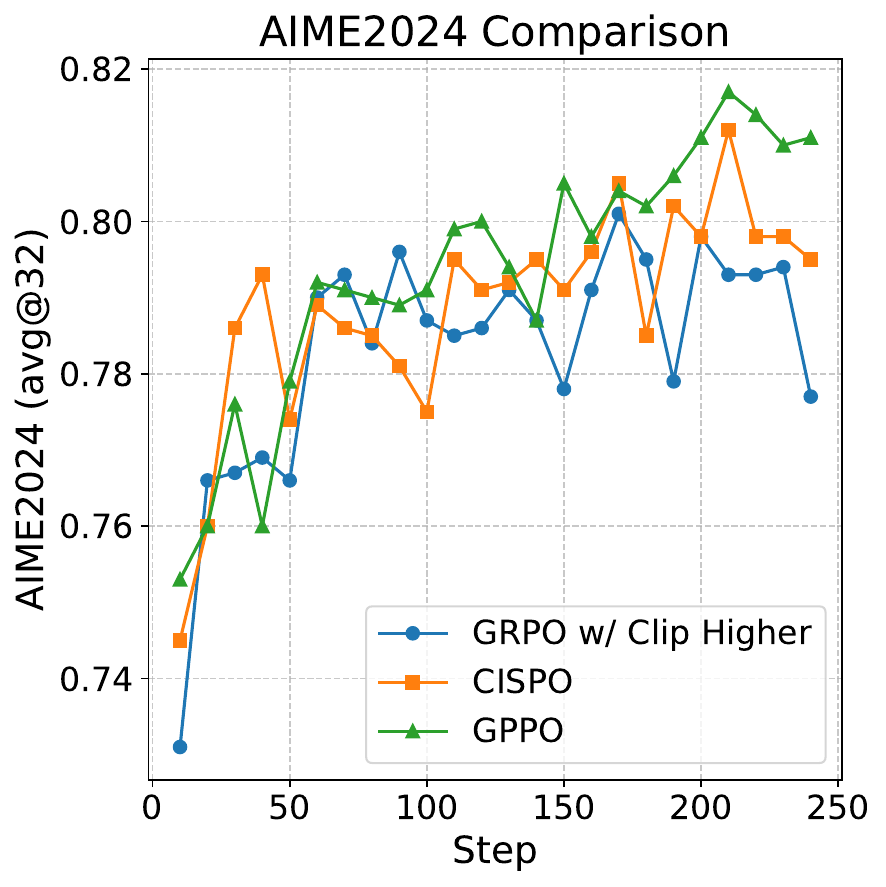}
        \label{fig:clip_aime24}
    } \hfill 
    \subfigure[LiveCodeBench]{
        \includegraphics[width=0.33\textwidth]{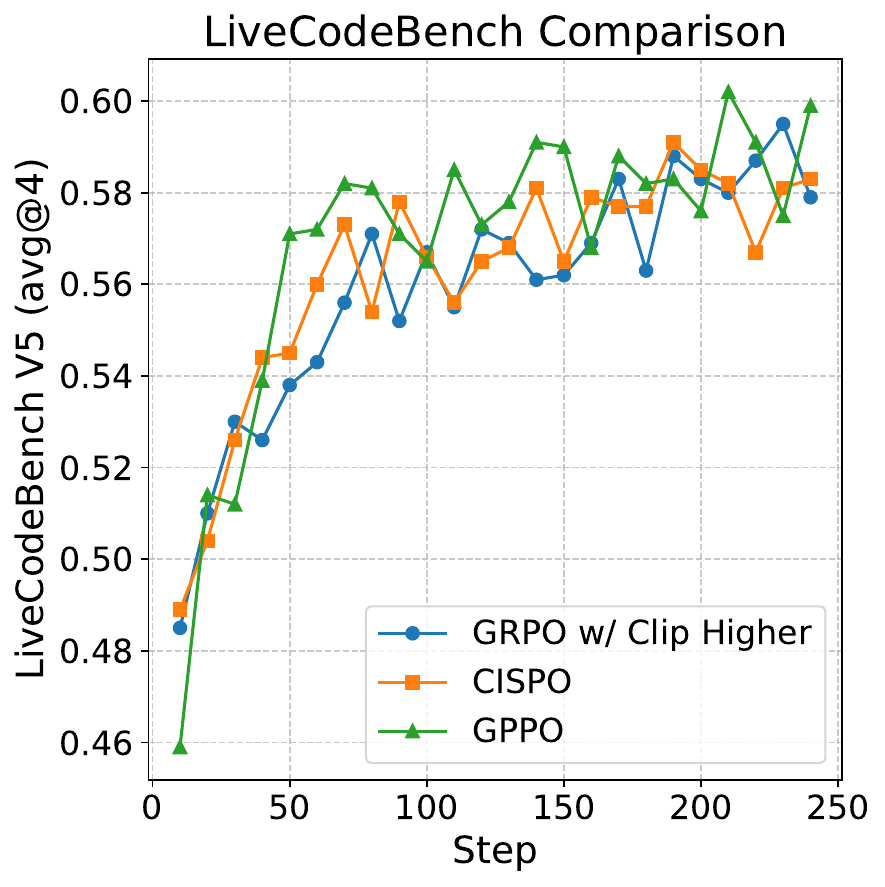}
        \label{fig:clip_lcb}
    } \hfill
    \subfigure[Grad Norm]{
        \includegraphics[width=0.33\textwidth]{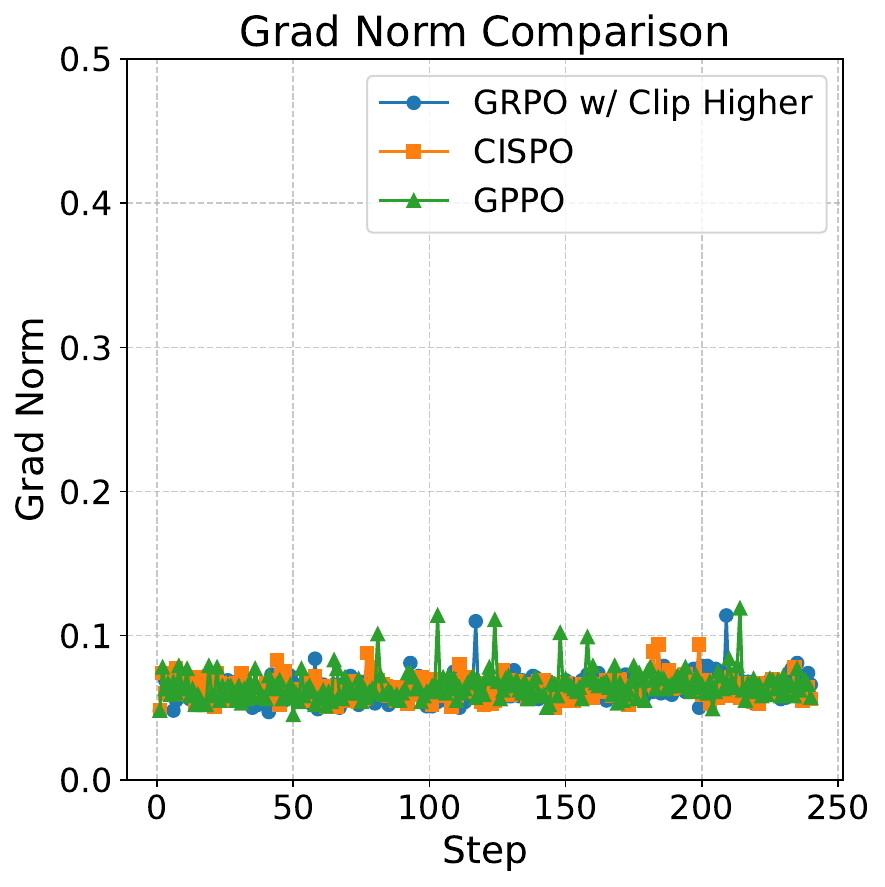}
        \label{fig:entropy}
    } 
    }
    \caption{Comparison of GPPO, GRPO w/ Clip Higher, and CISPO in mathematical RL training. Both methods are trained from an earlier long-CoT SFT checkpoint with a sequence length of 32K tokens. For GRPO, we use the Clip-Higher strategy from DAPO with the recommended $\epsilon_h = 0.28$.}
    \label{fig:clip_impact}
\end{figure*}

\subsubsection{Effectiveness of Gradient-Preserving Clipping Policy Optimization}
In this section, we examine the effects of GPPO on RL training. Figure \ref{fig:clip_aime24} and \ref{fig:clip_lcb} present the evaluation results on AIME2024 and LiveCodeBench V5. Compared to the traditional clipping method (\textit{GRPO w/ Clip-Higher}), our approach achieves superior performance on both benchmarks. We attribute these gains to two key factors:
\begin{itemize}[leftmargin=*]
    \item \textbf{Better utilization of high-entropy tokens.} High-entropy tokens often correspond to critical decision points in the policy. Traditional clipping methods may discard their gradients due to large importance sampling ratios, limiting the model’s exploration capability. Although DAPO's clip-higher method alleviates this problem, it does not fundamentally solve it, as tokens exceeding the boundary are still directly clipped. In contrast, our method preserves the gradients of these tokens and constrains them within a bounded range, updating model parameters in a gentle manner to ensure the model maintains its exploration capability at all times;
    \item \textbf{Accelerated convergence of negative samples.} Unlike traditional clipping which cuts off gradients from suboptimal trajectories that fall below the lower clipping bound, our method effectively leverages these negative samples. This prevents the model from repeatedly sampling similar negative trajectories, thereby enhancing its ability for rapid trial and error.
\end{itemize}

We further observe that CISPO, a concurrent work, achieves notable improvements on AIME2024 over the traditional clipping method, reinforcing the effectiveness of backpropagating gradients from clipped tokens. However, when directly comparing GPPO with CISPO, GPPO not only delivers higher stability in benchmark evaluations but also converges more effectively on both AIME2024 and LiveCodeBench V5.
We attribute this advantage to GPPO’s pessimistic update strategy.
Due to the $\min(\cdot)$ operation in GPPO, when the advantage \(\hat{A}_t\) is positive, GPPO suppresses overly optimistic updates, an effect that is absent in CISPO’s REINFORCE-based formulation. In contrast, when the advantage is negative, GPPO imposes no constraints on these updates, whereas CISPO limits them.
This pessimistic update mechanism helps GPPO maintain clearer optimization signals and leads to more stable policy training.

It is worth noting that, as shown in Figure \ref{fig:entropy}, all three methods maintained stable training, with most gradient norms remaining between 0.05 and 0.08. This indicates that decoupling gradient flow from clipping constraints not only improves learning ability but also preserves robustness advantages.

\subsubsection{Impact of SFT loss}

\begin{table}[!t]
\centering
\scalebox{1}{
\begin{tabular}{lcc}
\toprule
  & \textbf{AIME2024}   & \textbf{AIME2025}  \\
\midrule
Starting Checkpoint & 73.51 & 64.62 \\
\midrule
$\alpha=0$  & 78.13 & 68.02 \\
$\alpha=0.05$  & 77.81 & 69.48 \\
$\alpha=0.1$  & \textbf{79.37} & \textbf{70.42} \\
$\alpha=0.2$  & 78.44 & 69.58 \\
\bottomrule
\end{tabular}
}
\caption{Performance on AIME2024 (avg@32) and AIME2025 (avg@32) with varying $\alpha$ values when training with SFT loss. Here, $\alpha=0$ corresponds to the standard GRPO loss. All models are trained for 100 steps on mathematical RL data with a sequence length of 32K tokens.}
\label{alpha}
\end{table}

In this section, we aim to investigate the impact of different $\alpha$ values (the hyperparameter in Eq. \ref{loss:eq}). As shown in Table \ref{alpha}, most cases with $\alpha>0$ achieve better performance than $\alpha=0$, except for $\alpha=0.05$ which shows slight degradation on AIME2024. These results demonstrate that incorporating moderate SFT supervision can effectively guide the policy to produce higher-quality responses, benefiting model performance. The optimal performance is achieved when $\alpha$ equals 0.1. 

We posit that an appropriate $\alpha$ value enhances the utilization efficiency of positive examples during training. Simultaneously, SFT supervision acts as a policy regularizer, constraining the output distribution to remain within plausible bounds. However, excessive $\alpha$ disproportionately amplifies gradients from limited positive SFT samples, causing the policy to overfit these specific instances. This leads to localized overfitting and suppresses two critical capacities: negative learning from suboptimal trajectories, and autonomous exploration of novel solution spaces.

%



\begin{figure*}
    \centering
    \scalebox{0.90}{
    \subfigure[LiveCodeBench V5]{
        \includegraphics[width=0.4\textwidth]{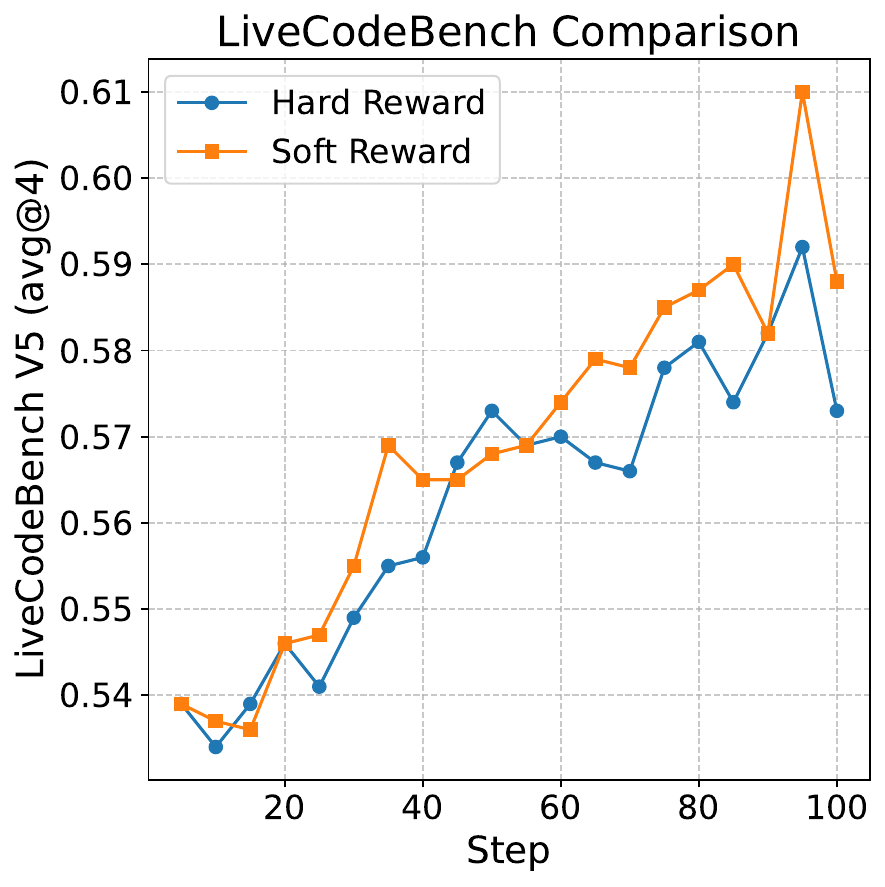}
        \label{fig:soft_hard_subfig1}
    } \hfill 
    \subfigure[Rewards]{
        \includegraphics[width=0.4\textwidth]{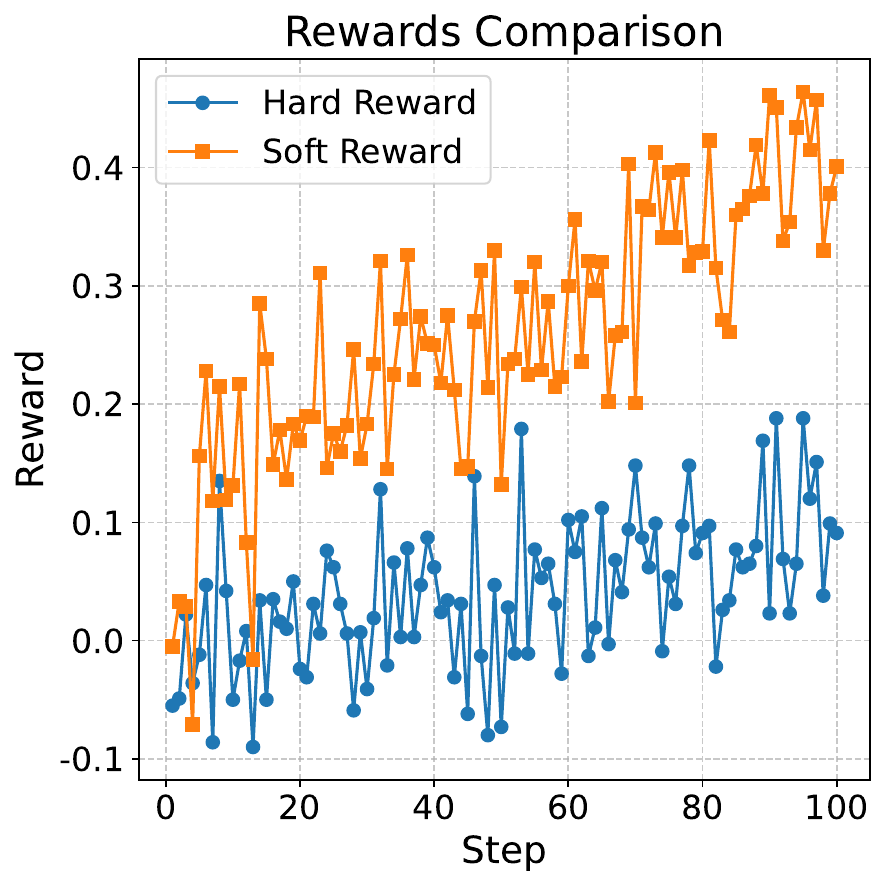}
        \label{fig:soft_hard_subfig2}
    }
    }
    \caption{Comparison between soft reward and hard reward strategies in code RL. Models are initialized from an early long CoT SFT checkpoint and trained for 100 steps with a sequence length of 16K tokens. In the soft reward setting, the reward equals the test case pass rate; in the hard reward setting, a positive reward is given only if all test cases pass and a negative reward otherwise.}
    \label{fig:overall_soft_hard}
\end{figure*}

\subsubsection{Soft Reward vs. Hard Reward}

The comparison between soft and hard reward strategies reveals clear advantages for the soft reward approach. As shown in Figures~\ref{fig:soft_hard_subfig1} and \ref{fig:soft_hard_subfig2}, assigning rewards proportional to the test case pass rate consistently yields higher average rewards than the hard reward baseline. This dense supervision allows the model to learn from partially correct outputs, alleviating reward sparsity and providing richer training signals.

Beyond improving average rewards, the soft reward also reduces reward variance, a crucial factor for stable policy optimization. In the hard reward setting, the binary nature of the reward leads to large fluctuations, particularly in the early stages when fully correct solutions are rare, thereby increasing the variance of policy gradient estimates and introducing noise into training. In contrast, the more continuous and structured feedback from soft reward stabilizes learning and guides the model toward better behaviors with more consistent gradients.

These improvements in both reward density and stability ultimately enhance downstream performance. As shown in Figure~\ref{fig:soft_hard_subfig1}, the model trained with soft reward achieves a LiveCodeBench V5 score of 61.0, outperforming the 59.2 achieved with hard reward. This 1.8-point gain confirms that addressing reward sparsity and variance not only improves learning dynamics, but also boosts generalization in code RL.

\begin{figure*}
    \centering
    \scalebox{0.90}{
    \includegraphics[width=0.4\textwidth]{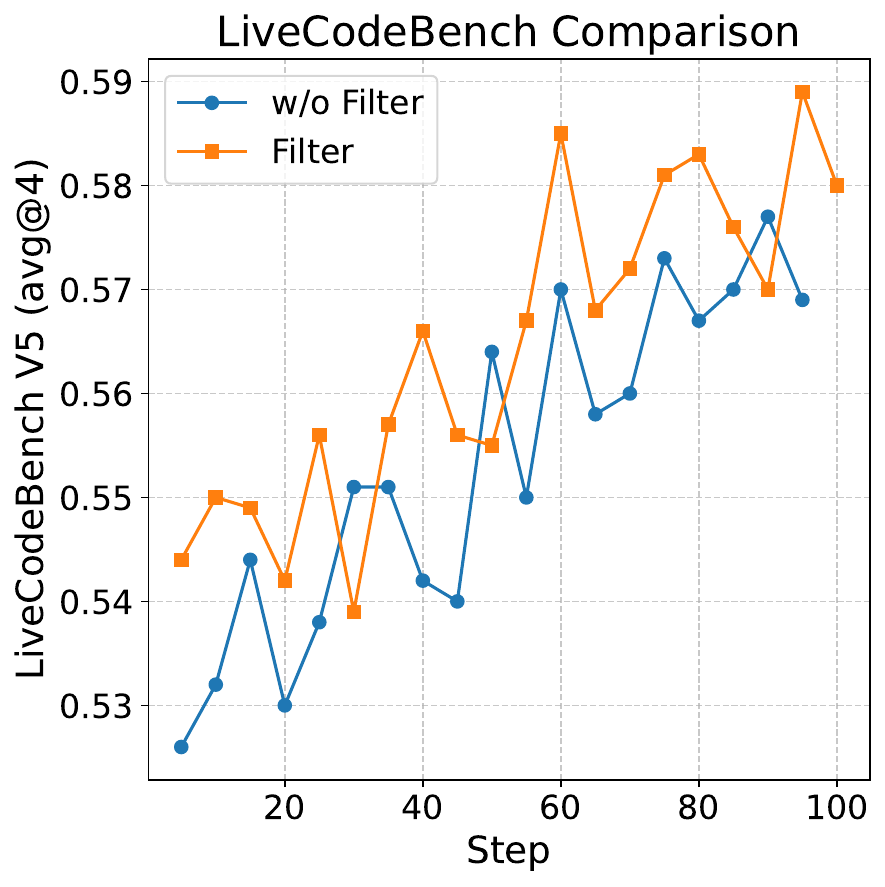}
    }
    \caption{Comparison of code RL performance on LiveCodeBench V5 (avg@4) using filtered versus unfiltered data. The filtering criterion retains prompts with estimated $pass@16 \geq 0.5$, based on 16 completions generated by DeepSeek-R1-0120. Models are initialized from an early-stage SFT checkpoint and trained for 100 steps with a sequence length of 16k tokens. \textit{Filter} denotes RL results with filtered data, while \textit{w/o Filter} represents results with the original unfiltered dataset.}
    \label{fig:code_filter}
\end{figure*}


\begin{figure*}
    \centering
    \scalebox{0.95}{
    \subfigure[AIME2024]{
        \includegraphics[width=0.4\textwidth]{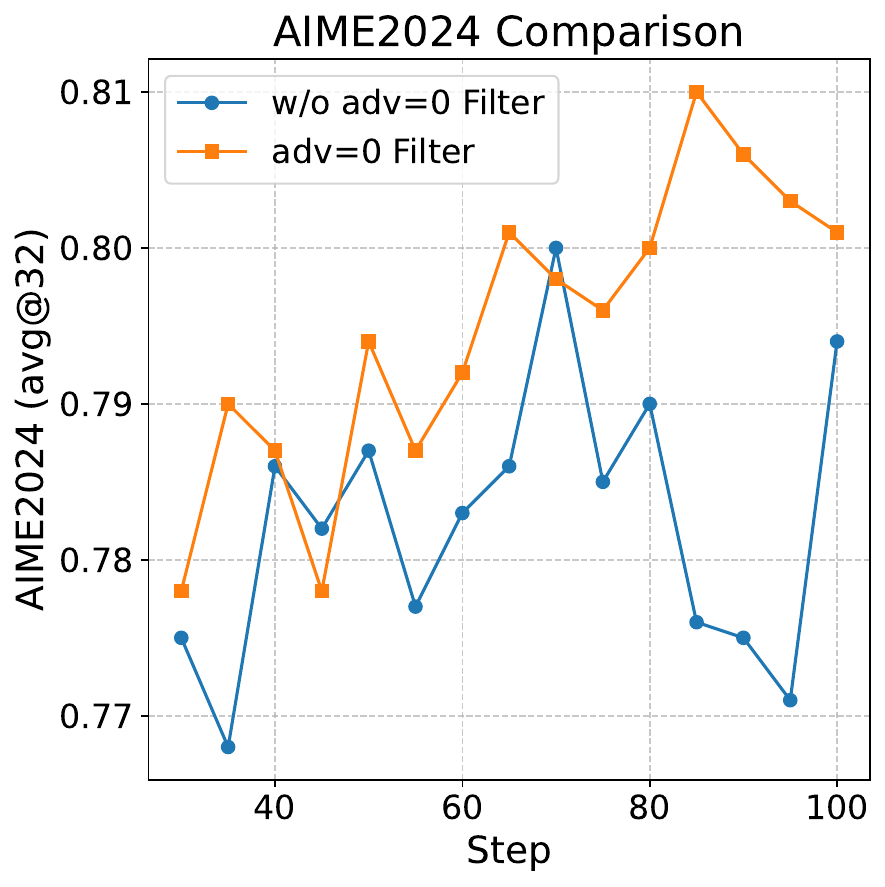}
        \label{fig:adv_aime}
    } \hfill 
    \subfigure[Rewards]{
        \includegraphics[width=0.4\textwidth]{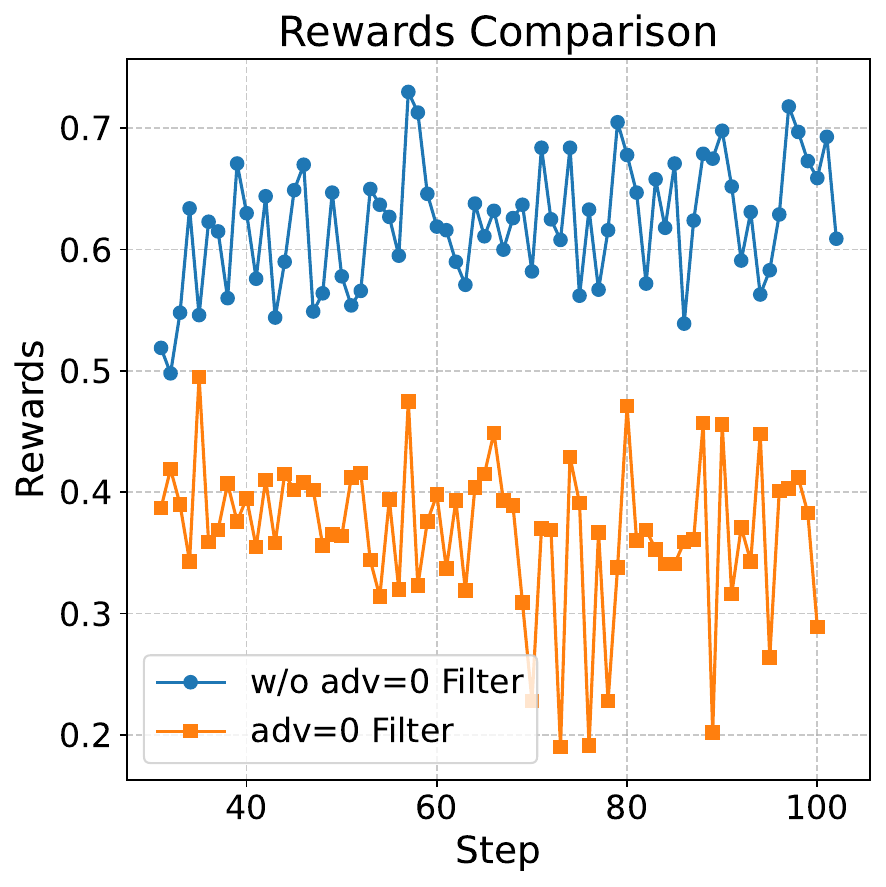}
        \label{fig:adv_rewards}
    }
    }
    \caption{Comparison of math RL performance on AIME2024 (avg@32) with and without filtering of zero-advantage groups. Experiments start from an intermediate RL checkpoint, training with a sequence length of 32K tokens. \textit{adv=0 Filter} denotes RL training excluding groups where all responses have zero advantage, while \textit{w/o adv=0 Filter} retains all sampled results regardless of advantage values. Notably, to maintain experimental rigor and ensure a fair comparison, we deliberately avoid supplementing the training data through dynamic sampling methods like those proposed by \citet{DBLP:journals/corr/abs-2503-14476}.}
    \label{fig:overall_adv_fig}
\end{figure*}

\subsubsection{Improving Code RL via Test Case Filtering}

We observe that certain test cases in open-source code datasets contain flaws that can cause correct code to fail execution. To mitigate the negative impact of such problematic cases during code RL training, we adopt a data filtering strategy based on estimated $pass@16$ scores. Applying this filtering to the DeepScaleR dataset and training for 100 RL steps, we find that models trained on filtered data consistently outperform those trained on unfiltered data in terms of final performance on LiveCodeBench V5, as illustrated in Figure \ref{fig:code_filter}. Moreover, the filtered-data models exhibit a more stable and steady upward trend throughout training. These improvements indicate that test case filtering is more effective because it reduces false negatives where correct completions are unfairly penalized due to flawed test cases. This results in a more reliable reinforcement learning signal and enables more efficient policy learning.


\subsubsection{Impact of Zero-Advantage Sample Filtering}


In GRPO, when all responses generated from a given prompt are either entirely correct or entirely incorrect, the advantage for that group evaluates to zero. This raises an important question about the role of such zero-advantage groups in reinforcement learning optimization. To investigate this empirically, we examine the effects of filtering out these zero-advantage groups during training.

The results, as illustrated in Figure \ref{fig:adv_aime}, reveal several findings. The filtered configuration demonstrates consistent and steady improvement on the AIME2024 benchmark throughout the RL training process. This stability suggests that the removal of zero-advantage samples helps maintain clear optimization signals. In contrast, while the unfiltered setting shows stable growth in reward values, its AIME2024 performance fluctuates significantly. This discrepancy indicates that while reward optimization may appear unaffected, the model's generalization capability becomes less predictable when zero-advantage samples are included.

The underlying mechanism for this behavior can be explained through the lens of policy gradient optimization. Zero-advantage samples inherently produce vanishing policy gradients, providing no meaningful learning signal. More critically, their presence in large quantities effectively dilutes the contribution of non-zero advantage samples, creating ambiguity in the optimization direction. This not only slows down learning but may also introduce noise into the training process, potentially explaining the observed instability in generalization performance. The zero-advantage sample filtering approach appears to mitigate these issues by focusing the model's learning on samples that provide clearer, more actionable feedback.

\section{Conclusions}
Klear-Reasoner advances reasoning capabilities in mathematics and programming by integrating long Chain-of-Thought supervised fine-tuning with Gradient-Preserving Clipping Policy Optimization. The quality-centric SFT strategy ensures consistent learning of accurate reasoning patterns, while GPPO addresses the limitations of traditional clipping by retaining gradient information from all tokens, enabling both stable policy updates and effective exploration. Complementary techniques, such as soft reward design, data filtering, and balanced SFT supervision, further enhance reinforcement learning efficiency. Evaluations on multiple challenging benchmarks show that Klear-Reasoner consistently matches or surpasses state-of-the-art models of comparable scale. These results demonstrate that principled data curation, targeted SFT, and carefully designed RL optimization can jointly deliver substantial improvements in long-form reasoning performance.

\bibliography{iclr2025_conference}

\begin{thebibliography}{28}
\providecommand{\natexlab}[1]{#1}
\providecommand{\url}[1]{\texttt{#1}}
\expandafter\ifx\csname urlstyle\endcsname\relax
  \providecommand{\doi}[1]{doi: #1}\else
  \providecommand{\doi}{doi: \begingroup \urlstyle{rm}\Url}\fi

\bibitem[Ahmad et~al.(2025{\natexlab{a}})Ahmad, Majumdar, Ficek, Narenthiran, Samadi, Huang, Jain, Noroozi, and Ginsburg]{ahmad2025opencodereasoningiisimpletesttime}
Wasi~Uddin Ahmad, Somshubra Majumdar, Aleksander Ficek, Sean Narenthiran, Mehrzad Samadi, Jocelyn Huang, Siddhartha Jain, Vahid Noroozi, and Boris Ginsburg.
\newblock {OpenCodeReasoning-II: A Simple Test Time Scaling Approach via Self-Critique}, 2025{\natexlab{a}}.
\newblock URL \url{https://arxiv.org/abs/2507.09075}.

\bibitem[Ahmad et~al.(2025{\natexlab{b}})Ahmad, Narenthiran, Majumdar, Ficek, Jain, Huang, Noroozi, and Ginsburg]{DBLP:journals/corr/abs-2504-01943}
Wasi~Uddin Ahmad, Sean Narenthiran, Somshubra Majumdar, Aleksander Ficek, Siddhartha Jain, Jocelyn Huang, Vahid Noroozi, and Boris Ginsburg.
\newblock Opencodereasoning: Advancing data distillation for competitive coding.
\newblock \emph{CoRR}, abs/2504.01943, 2025{\natexlab{b}}.
\newblock \doi{10.48550/ARXIV.2504.01943}.
\newblock URL \url{https://doi.org/10.48550/arXiv.2504.01943}.

\bibitem[An et~al.(2025)An, Xie, Li, Li, Zhang, Gong, Zhong, Xu, Qiu, Wang, and Kong]{Polaris2025}
Chenxin An, Zhihui Xie, Xiaonan Li, Lei Li, Jun Zhang, Shansan Gong, Ming Zhong, Jingjing Xu, Xipeng Qiu, Mingxuan Wang, and Lingpeng Kong.
\newblock Polaris: A post-training recipe for scaling reinforcement learning on advanced reasoning models, 2025.
\newblock URL \url{https://hkunlp.github.io/blog/2025/Polaris}.

\bibitem[Chen et~al.(2025{\natexlab{a}})Chen, Li, Gong, Jiang, Fei, Yang, Shan, Yu, Wang, Zhu, Xiao, Du, Zhang, Qiao, Zhang, Du, Guo, Chen, Ding, Sun, Li, Jiao, Zhou, Zhang, Ding, Sun, Feng, Cai, Zhu, Sun, Zhuang, Cai, Song, Zhu, Li, Tian, Liu, Xu, Yan, Liu, He, Feng, Yang, Xiao, Han, Wang, Yu, Feng, Li, Zheng, Du, Yang, Zeng, Yu, Tao, Chi, Zhang, Lin, Hu, Di, Gao, Li, Zhao, Ren, Xu, Li, Wang, Tian, Leng, Chen, Chen, Shi, Weng, Guan, Yu, Li, Zhu, Li, Cai, Liang, Cheng, Kong, Li, Chen, Song, Luo, Su, Li, Han, Hou, Lu, Zou, Shen, Gong, Ma, Wang, Shi, Zhong, and Duan]{DBLP:journals/corr/abs-2506-13585}
Aili Chen, Aonian Li, Bangwei Gong, Binyang Jiang, Bo~Fei, Bo~Yang, Boji Shan, Changqing Yu, Chao Wang, Cheng Zhu, Chengjun Xiao, Chengyu Du, Chi Zhang, Chu Qiao, Chunhao Zhang, Chunhui Du, Congchao Guo, Da~Chen, Deming Ding, Dianjun Sun, Dong Li, Enwei Jiao, Haigang Zhou, Haimo Zhang, Han Ding, Haohai Sun, Haoyu Feng, Huaiguang Cai, Haichao Zhu, Jian Sun, Jiaqi Zhuang, Jiaren Cai, Jiayuan Song, Jin Zhu, Jingyang Li, Jinhao Tian, Jinli Liu, Junhao Xu, Junjie Yan, Junteng Liu, Junxian He, Kaiyi Feng, Ke~Yang, Kecheng Xiao, Le~Han, Leyang Wang, Lianfei Yu, Liheng Feng, Lin Li, Lin Zheng, Linge Du, Lingyu Yang, Lunbin Zeng, Minghui Yu, Mingliang Tao, Mingyuan Chi, Mozhi Zhang, Mujie Lin, Nan Hu, Nongyu Di, Peng Gao, Pengfei Li, Pengyu Zhao, Qibing Ren, Qidi Xu, Qile Li, Qin Wang, Rong Tian, Ruitao Leng, Shaoxiang Chen, Shaoyu Chen, Shengmin Shi, Shitong Weng, Shuchang Guan, Shuqi Yu, Sichen Li, Songquan Zhu, Tengfei Li, Tianchi Cai, Tianrun Liang, Weiyu Cheng, Weize Kong, Wenkai Li, Xiancai Chen, Xiangjun Song,
  Xiao Luo, Xiao Su, Xiaobo Li, Xiaodong Han, Xinzhu Hou, Xuan Lu, Xun Zou, Xuyang Shen, Yan Gong, Yan Ma, Yang Wang, Yiqi Shi, Yiran Zhong, and Yonghong Duan.
\newblock Minimax-m1: Scaling test-time compute efficiently with lightning attention.
\newblock \emph{CoRR}, abs/2506.13585, 2025{\natexlab{a}}.
\newblock \doi{10.48550/ARXIV.2506.13585}.
\newblock URL \url{https://doi.org/10.48550/arXiv.2506.13585}.

\bibitem[Chen et~al.(2025{\natexlab{b}})Chen, Yang, Liu, Lee, Xu, Shoeybi, Catanzaro, and Ping]{DBLP:journals/corr/abs-2505-16400}
Yang Chen, Zhuolin Yang, Zihan Liu, Chankyu Lee, Peng Xu, Mohammad Shoeybi, Bryan Catanzaro, and Wei Ping.
\newblock Acereason-nemotron: Advancing math and code reasoning through reinforcement learning.
\newblock \emph{CoRR}, abs/2505.16400, 2025{\natexlab{b}}.
\newblock \doi{10.48550/ARXIV.2505.16400}.
\newblock URL \url{https://doi.org/10.48550/arXiv.2505.16400}.

\bibitem[DeepSeek{-}AI et~al.(2025)DeepSeek{-}AI, Guo, Yang, Zhang, Song, Zhang, Xu, Zhu, Ma, Wang, Bi, Zhang, Yu, Wu, Wu, Gou, Shao, Li, Gao, Liu, Xue, Wang, Wu, Feng, Lu, Zhao, Deng, Zhang, Ruan, Dai, Chen, Ji, Li, Lin, Dai, Luo, Hao, Chen, Li, Zhang, Bao, Xu, Wang, Ding, Xin, Gao, Qu, Li, Guo, Li, Wang, Chen, Yuan, Qiu, Li, Cai, Ni, Liang, Chen, Dong, Hu, Gao, Guan, Huang, Yu, Wang, Zhang, Zhao, Wang, Zhang, Xu, Xia, Zhang, Zhang, Tang, Li, Wang, Li, Tian, Huang, Zhang, Wang, Chen, Du, Ge, Zhang, Pan, Wang, Chen, Jin, Chen, Lu, Zhou, Chen, Ye, Wang, Yu, Zhou, Pan, and Li]{DBLP:journals/corr/abs-2501-12948}
DeepSeek{-}AI, Daya Guo, Dejian Yang, Haowei Zhang, Junxiao Song, Ruoyu Zhang, Runxin Xu, Qihao Zhu, Shirong Ma, Peiyi Wang, Xiao Bi, Xiaokang Zhang, Xingkai Yu, Yu~Wu, Z.~F. Wu, Zhibin Gou, Zhihong Shao, Zhuoshu Li, Ziyi Gao, Aixin Liu, Bing Xue, Bingxuan Wang, Bochao Wu, Bei Feng, Chengda Lu, Chenggang Zhao, Chengqi Deng, Chenyu Zhang, Chong Ruan, Damai Dai, Deli Chen, Dongjie Ji, Erhang Li, Fangyun Lin, Fucong Dai, Fuli Luo, Guangbo Hao, Guanting Chen, Guowei Li, H.~Zhang, Han Bao, Hanwei Xu, Haocheng Wang, Honghui Ding, Huajian Xin, Huazuo Gao, Hui Qu, Hui Li, Jianzhong Guo, Jiashi Li, Jiawei Wang, Jingchang Chen, Jingyang Yuan, Junjie Qiu, Junlong Li, J.~L. Cai, Jiaqi Ni, Jian Liang, Jin Chen, Kai Dong, Kai Hu, Kaige Gao, Kang Guan, Kexin Huang, Kuai Yu, Lean Wang, Lecong Zhang, Liang Zhao, Litong Wang, Liyue Zhang, Lei Xu, Leyi Xia, Mingchuan Zhang, Minghua Zhang, Minghui Tang, Meng Li, Miaojun Wang, Mingming Li, Ning Tian, Panpan Huang, Peng Zhang, Qiancheng Wang, Qinyu Chen, Qiushi Du, Ruiqi Ge,
  Ruisong Zhang, Ruizhe Pan, Runji Wang, R.~J. Chen, R.~L. Jin, Ruyi Chen, Shanghao Lu, Shangyan Zhou, Shanhuang Chen, Shengfeng Ye, Shiyu Wang, Shuiping Yu, Shunfeng Zhou, Shuting Pan, and S.~S. Li.
\newblock Deepseek-r1: Incentivizing reasoning capability in llms via reinforcement learning.
\newblock \emph{CoRR}, abs/2501.12948, 2025.
\newblock \doi{10.48550/ARXIV.2501.12948}.
\newblock URL \url{https://doi.org/10.48550/arXiv.2501.12948}.

\bibitem[Fu et~al.(2025)Fu, Gao, Shen, Zhu, Mei, He, Xu, Wei, Mei, Wang, Yang, Yuan, and Wu]{fu2025areal}
Wei Fu, Jiaxuan Gao, Xujie Shen, Chen Zhu, Zhiyu Mei, Chuyi He, Shusheng Xu, Guo Wei, Jun Mei, Jiashu Wang, Tongkai Yang, Binhang Yuan, and Yi~Wu.
\newblock Areal: A large-scale asynchronous reinforcement learning system for language reasoning, 2025.
\newblock URL \url{https://arxiv.org/abs/2505.24298}.

\bibitem[Guha et~al.(2025)Guha, Marten, Keh, Raoof, Smyrnis, Bansal, Nezhurina, Mercat, Vu, Sprague, Suvarna, Feuer, Chen, Khan, Frankel, Grover, Choi, Muennighoff, Su, Zhao, Yang, Pimpalgaonkar, Sharma, Ji, Deng, Pratt, Ramanujan, Saad{-}Falcon, Li, Dave, Albalak, Arora, Wulfe, Hegde, Durrett, Oh, Bansal, Gabriel, Grover, Chang, Shankar, Gokaslan, Merrill, Hashimoto, Choi, Jitsev, Heckel, Sathiamoorthy, Dimakis, and Schmidt]{DBLP:journals/corr/abs-2506-04178}
Etash~Kumar Guha, Ryan Marten, Sedrick Keh, Negin Raoof, Georgios Smyrnis, Hritik Bansal, Marianna Nezhurina, Jean Mercat, Trung Vu, Zayne Sprague, Ashima Suvarna, Benjamin Feuer, Liangyu Chen, Zaid Khan, Eric Frankel, Sachin Grover, Caroline Choi, Niklas Muennighoff, Shiye Su, Wanjia Zhao, John Yang, Shreyas Pimpalgaonkar, Kartik Sharma, Charlie~Cheng{-}Jie Ji, Yichuan Deng, Sarah~M. Pratt, Vivek Ramanujan, Jon Saad{-}Falcon, Jeffrey Li, Achal Dave, Alon Albalak, Kushal Arora, Blake Wulfe, Chinmay Hegde, Greg Durrett, Sewoong Oh, Mohit Bansal, Saadia Gabriel, Aditya Grover, Kai{-}Wei Chang, Vaishaal Shankar, Aaron Gokaslan, Mike~A. Merrill, Tatsunori Hashimoto, Yejin Choi, Jenia Jitsev, Reinhard Heckel, Maheswaran Sathiamoorthy, Alexandros~G. Dimakis, and Ludwig Schmidt.
\newblock Openthoughts: Data recipes for reasoning models.
\newblock \emph{CoRR}, abs/2506.04178, 2025.
\newblock \doi{10.48550/ARXIV.2506.04178}.
\newblock URL \url{https://doi.org/10.48550/arXiv.2506.04178}.

\bibitem[He et~al.(2025)He, Liu, Liu, Yan, Wang, Cheng, Zhang, Zhang, Xu, Shen, Li, Zeng, Wei, Cheng, An, Liu, and Zhou]{DBLP:journals/corr/abs-2505-22312}
Jujie He, Jiacai Liu, Chris~Yuhao Liu, Rui Yan, Chaojie Wang, Peng Cheng, Xiaoyu Zhang, Fuxiang Zhang, Jiacheng Xu, Wei Shen, Siyuan Li, Liang Zeng, Tianwen Wei, Cheng Cheng, Bo~An, Yang Liu, and Yahui Zhou.
\newblock Skywork open reasoner 1 technical report.
\newblock \emph{CoRR}, abs/2505.22312, 2025.
\newblock \doi{10.48550/ARXIV.2505.22312}.
\newblock URL \url{https://doi.org/10.48550/arXiv.2505.22312}.

\bibitem[Hendrycks et~al.(2021)Hendrycks, Basart, Kadavath, Mazeika, Arora, Guo, Burns, Puranik, He, Song, and Steinhardt]{hendrycksapps2021}
Dan Hendrycks, Steven Basart, Saurav Kadavath, Mantas Mazeika, Akul Arora, Ethan Guo, Collin Burns, Samir Puranik, Horace He, Dawn Song, and Jacob Steinhardt.
\newblock Measuring coding challenge competence with apps.
\newblock \emph{NeurIPS}, 2021.

\bibitem[Jaech et~al.(2024)Jaech, Kalai, Lerer, Richardson, El{-}Kishky, Low, Helyar, Madry, Beutel, Carney, Iftimie, Karpenko, Passos, Neitz, Prokofiev, Wei, Tam, Bennett, Kumar, Saraiva, Vallone, Duberstein, Kondrich, Mishchenko, Applebaum, Jiang, Nair, Zoph, Ghorbani, Rossen, Sokolowsky, Barak, McGrew, Minaiev, Hao, Baker, Houghton, McKinzie, Eastman, Lugaresi, Bassin, Hudson, Li, de~Bourcy, Voss, Shen, Zhang, Koch, Orsinger, Hesse, Fischer, Chan, Roberts, Kappler, Levy, Selsam, Dohan, Farhi, Mely, Robinson, Tsipras, Li, Oprica, Freeman, Zhang, Wong, Proehl, Cheung, Mitchell, Wallace, Ritter, Mays, Wang, Such, Raso, Leoni, Tsimpourlas, Song, von Lohmann, Sulit, Salmon, Parascandolo, Chabot, Zhao, Brockman, Leclerc, Salman, Bao, Sheng, Andrin, Bagherinezhad, Ren, Lightman, Chung, Kivlichan, O'Connell, Osband, Gilaberte, and Akkaya]{DBLP:journals/corr/abs-2412-16720}
Aaron Jaech, Adam Kalai, Adam Lerer, Adam Richardson, Ahmed El{-}Kishky, Aiden Low, Alec Helyar, Aleksander Madry, Alex Beutel, Alex Carney, Alex Iftimie, Alex Karpenko, Alex~Tachard Passos, Alexander Neitz, Alexander Prokofiev, Alexander Wei, Allison Tam, Ally Bennett, Ananya Kumar, Andre Saraiva, Andrea Vallone, Andrew Duberstein, Andrew Kondrich, Andrey Mishchenko, Andy Applebaum, Angela Jiang, Ashvin Nair, Barret Zoph, Behrooz Ghorbani, Ben Rossen, Benjamin Sokolowsky, Boaz Barak, Bob McGrew, Borys Minaiev, Botao Hao, Bowen Baker, Brandon Houghton, Brandon McKinzie, Brydon Eastman, Camillo Lugaresi, Cary Bassin, Cary Hudson, Chak~Ming Li, Charles de~Bourcy, Chelsea Voss, Chen Shen, Chong Zhang, Chris Koch, Chris Orsinger, Christopher Hesse, Claudia Fischer, Clive Chan, Dan Roberts, Daniel Kappler, Daniel Levy, Daniel Selsam, David Dohan, David Farhi, David Mely, David Robinson, Dimitris Tsipras, Doug Li, Dragos Oprica, Eben Freeman, Eddie Zhang, Edmund Wong, Elizabeth Proehl, Enoch Cheung, Eric Mitchell,
  Eric Wallace, Erik Ritter, Evan Mays, Fan Wang, Felipe~Petroski Such, Filippo Raso, Florencia Leoni, Foivos Tsimpourlas, Francis Song, Fred von Lohmann, Freddie Sulit, Geoff Salmon, Giambattista Parascandolo, Gildas Chabot, Grace Zhao, Greg Brockman, Guillaume Leclerc, Hadi Salman, Haiming Bao, Hao Sheng, Hart Andrin, Hessam Bagherinezhad, Hongyu Ren, Hunter Lightman, Hyung~Won Chung, Ian Kivlichan, Ian O'Connell, Ian Osband, Ignasi~Clavera Gilaberte, and Ilge Akkaya.
\newblock Openai o1 system card.
\newblock \emph{CoRR}, abs/2412.16720, 2024.
\newblock \doi{10.48550/ARXIV.2412.16720}.
\newblock URL \url{https://doi.org/10.48550/arXiv.2412.16720}.

\bibitem[LI et~al.(2024)LI, Beeching, Tunstall, Lipkin, Soletskyi, Huang, Rasul, Yu, Jiang, Shen, Qin, Dong, Zhou, Fleureau, Lample, and Polu]{numina_math_datasets}
Jia LI, Edward Beeching, Lewis Tunstall, Ben Lipkin, Roman Soletskyi, Shengyi~Costa Huang, Kashif Rasul, Longhui Yu, Albert Jiang, Ziju Shen, Zihan Qin, Bin Dong, Li~Zhou, Yann Fleureau, Guillaume Lample, and Stanislas Polu.
\newblock Numinamath.
\newblock \url{[https://huggingface.co/AI-MO/NuminaMath-CoT](https://github.com/project-numina/aimo-progress-prize/blob/main/report/numina_dataset.pdf)}, 2024.

\bibitem[Li et~al.(2023)Li, Fu, Zhang, Huang, Sun, Lyu, Liu, Jin, and Li]{li2023taco}
Rongao Li, Jie Fu, Bo-Wen Zhang, Tao Huang, Zhihong Sun, Chen Lyu, Guang Liu, Zhi Jin, and Ge~Li.
\newblock Taco: Topics in algorithmic code generation dataset.
\newblock \emph{arXiv preprint arXiv:2312.14852}, 2023.

\bibitem[Liu et~al.(2025)Liu, Yang, Chen, Lee, Shoeybi, Catanzaro, and Ping]{DBLP:journals/corr/abs-2506-13284}
Zihan Liu, Zhuolin Yang, Yang Chen, Chankyu Lee, Mohammad Shoeybi, Bryan Catanzaro, and Wei Ping.
\newblock Acereason-nemotron 1.1: Advancing math and code reasoning through {SFT} and {RL} synergy.
\newblock \emph{CoRR}, abs/2506.13284, 2025.
\newblock \doi{10.48550/ARXIV.2506.13284}.
\newblock URL \url{https://doi.org/10.48550/arXiv.2506.13284}.

\bibitem[Luo et~al.(2025{\natexlab{a}})Luo, Tan, Huang, Patel, Ariyak, Wu, Shi, Xin, Cai, Weber, Zhang, Li, Popa, and Stoica]{deepcoder2025}
Michael Luo, Sijun Tan, Roy Huang, Ameen Patel, Alpay Ariyak, Qingyang Wu, Xiaoxiang Shi, Rachel Xin, Colin Cai, Maurice Weber, Ce~Zhang, Li~Erran Li, Raluca~Ada Popa, and Ion Stoica.
\newblock Deepcoder: A fully open-source 14b coder at o3-mini level, 2025{\natexlab{a}}.
\newblock Notion Blog.

\bibitem[Luo et~al.(2025{\natexlab{b}})Luo, Tan, Wong, Shi, Tang, Roongta, Cai, Luo, Li, Popa, and Stoica]{deepScaler2025}
Michael Luo, Sijun Tan, Justin Wong, Xiaoxiang Shi, William~Y. Tang, Manan Roongta, Colin Cai, Jeffrey Luo, Li~Erran Li, Raluca~Ada Popa, and Ion Stoica.
\newblock Deepscaler: Surpassing o1-preview with a 1.5b model by scaling rl, 2025{\natexlab{b}}.
\newblock Notion Blog.

\bibitem[Penedo et~al.(2025)Penedo, Lozhkov, Kydlíček, Allal, Beeching, Lajarín, Gallouédec, Habib, Tunstall, and von Werra]{penedo2025codeforces}
Guilherme Penedo, Anton Lozhkov, Hynek Kydlíček, Loubna~Ben Allal, Edward Beeching, Agustín~Piqueres Lajarín, Quentin Gallouédec, Nathan Habib, Lewis Tunstall, and Leandro von Werra.
\newblock Codeforces.
\newblock \url{https://huggingface.co/datasets/open-r1/codeforces}, 2025.

\bibitem[Peng et~al.(2024)Peng, Quesnelle, Fan, and Shippole]{DBLP:conf/iclr/PengQFS24}
Bowen Peng, Jeffrey Quesnelle, Honglu Fan, and Enrico Shippole.
\newblock Yarn: Efficient context window extension of large language models.
\newblock In \emph{The Twelfth International Conference on Learning Representations, {ICLR} 2024, Vienna, Austria, May 7-11, 2024}. OpenReview.net, 2024.
\newblock URL \url{https://openreview.net/forum?id=wHBfxhZu1u}.

\bibitem[Schulman et~al.(2015)Schulman, Levine, Abbeel, Jordan, and Moritz]{DBLP:conf/icml/SchulmanLAJM15}
John Schulman, Sergey Levine, Pieter Abbeel, Michael~I. Jordan, and Philipp Moritz.
\newblock Trust region policy optimization.
\newblock In Francis~R. Bach and David~M. Blei (eds.), \emph{Proceedings of the 32nd International Conference on Machine Learning, {ICML} 2015, Lille, France, 6-11 July 2015}, volume~37 of \emph{{JMLR} Workshop and Conference Proceedings}, pp.\  1889--1897. JMLR.org, 2015.
\newblock URL \url{http://proceedings.mlr.press/v37/schulman15.html}.

\bibitem[Schulman et~al.(2016)Schulman, Moritz, Levine, Jordan, and Abbeel]{DBLP:journals/corr/SchulmanMLJA15}
John Schulman, Philipp Moritz, Sergey Levine, Michael~I. Jordan, and Pieter Abbeel.
\newblock High-dimensional continuous control using generalized advantage estimation.
\newblock In Yoshua Bengio and Yann LeCun (eds.), \emph{4th International Conference on Learning Representations, {ICLR} 2016, San Juan, Puerto Rico, May 2-4, 2016, Conference Track Proceedings}, 2016.
\newblock URL \url{http://arxiv.org/abs/1506.02438}.

\bibitem[Schulman et~al.(2017)Schulman, Wolski, Dhariwal, Radford, and Klimov]{DBLP:journals/corr/SchulmanWDRK17}
John Schulman, Filip Wolski, Prafulla Dhariwal, Alec Radford, and Oleg Klimov.
\newblock Proximal policy optimization algorithms.
\newblock \emph{CoRR}, abs/1707.06347, 2017.
\newblock URL \url{http://arxiv.org/abs/1707.06347}.

\bibitem[Shao et~al.(2024)Shao, Wang, Zhu, Xu, Song, Zhang, Li, Wu, and Guo]{DBLP:journals/corr/abs-2402-03300}
Zhihong Shao, Peiyi Wang, Qihao Zhu, Runxin Xu, Junxiao Song, Mingchuan Zhang, Y.~K. Li, Y.~Wu, and Daya Guo.
\newblock Deepseekmath: Pushing the limits of mathematical reasoning in open language models.
\newblock \emph{CoRR}, abs/2402.03300, 2024.
\newblock \doi{10.48550/ARXIV.2402.03300}.
\newblock URL \url{https://doi.org/10.48550/arXiv.2402.03300}.

\bibitem[Team et~al.(2025)Team, Hu, Chen, Zhao, Liu, Jin, Zhu, Dai, Luan, Guo, Liu, Wu, Mei, Zhou, Zhao, Xiong, Zhang, Xu, Liang, Jiang, Fu, Zheng, Gao, Cui, Wan, Zheng, Li, Yang, Ren, Yan, Wan, Feng, Zhao, Yang, Kong, Yang, Li, Wu, Liu, Xu, Zhang, Zhou, Huang, Zhang, Wang, and Wen]{DBLP:journals/corr/abs-2506-14731}
Ling Team, Bin Hu, Cai Chen, Deng Zhao, Ding Liu, Dingnan Jin, Feng Zhu, Hao Dai, Hongzhi Luan, Jia Guo, Jiaming Liu, Jiewei Wu, Jun Mei, Jun Zhou, Junbo Zhao, Junwu Xiong, Kaihong Zhang, Kuan Xu, Lei Liang, Liang Jiang, Liangcheng Fu, Longfei Zheng, Qiang Gao, Qing Cui, Quan Wan, Shaomian Zheng, Shuaicheng Li, Tongkai Yang, Wang Ren, Xiaodong Yan, Xiaopei Wan, Xiaoyun Feng, Xin Zhao, Xinxing Yang, Xinyu Kong, Xuemin Yang, Yang Li, Yingting Wu, Yongkang Liu, Zhankai Xu, Zhenduo Zhang, Zhenglei Zhou, Zhenyu Huang, Zhiqiang Zhang, Zihao Wang, and Zujie Wen.
\newblock Ring-lite: Scalable reasoning via c3po-stabilized reinforcement learning for llms.
\newblock \emph{CoRR}, abs/2506.14731, 2025.
\newblock \doi{10.48550/ARXIV.2506.14731}.
\newblock URL \url{https://doi.org/10.48550/arXiv.2506.14731}.

\bibitem[Wen et~al.(2025)Wen, Cai, Xiao, He, An, Duan, Du, Liu, Tang, Lv, Zou, Deng, Jia, and Zhang]{DBLP:journals/corr/abs-2503-10460}
Liang Wen, Yunke Cai, Fenrui Xiao, Xin He, Qi~An, Zhenyu Duan, Yimin Du, Junchen Liu, Lifu Tang, Xiaowei Lv, Haosheng Zou, Yongchao Deng, Shousheng Jia, and Xiangzheng Zhang.
\newblock Light-r1: Curriculum sft, {DPO} and {RL} for long {COT} from scratch and beyond.
\newblock \emph{CoRR}, abs/2503.10460, 2025.
\newblock \doi{10.48550/ARXIV.2503.10460}.
\newblock URL \url{https://doi.org/10.48550/arXiv.2503.10460}.

\bibitem[Xiaomi(2025)]{coreteam2025mimounlockingreasoningpotential}
LLM-Core-Team Xiaomi.
\newblock Mimo: Unlocking the reasoning potential of language model -- from pretraining to posttraining, 2025.
\newblock URL \url{https://arxiv.org/abs/2505.07608}.

\bibitem[Yang et~al.(2025)Yang, Li, Yang, Zhang, Hui, Zheng, Yu, Gao, Huang, Lv, Zheng, Liu, Zhou, Huang, Hu, Ge, Wei, Lin, Tang, Yang, Tu, Zhang, Yang, Yang, Zhou, Zhou, Lin, Dang, Bao, Yang, Yu, Deng, Li, Xue, Li, Zhang, Wang, Zhu, Men, Gao, Liu, Luo, Li, Tang, Yin, Ren, Wang, Zhang, Ren, Fan, Su, Zhang, Zhang, Wan, Liu, Wang, Cui, Zhang, Zhou, and Qiu]{DBLP:journals/corr/abs-2505-09388}
An~Yang, Anfeng Li, Baosong Yang, Beichen Zhang, Binyuan Hui, Bo~Zheng, Bowen Yu, Chang Gao, Chengen Huang, Chenxu Lv, Chujie Zheng, Dayiheng Liu, Fan Zhou, Fei Huang, Feng Hu, Hao Ge, Haoran Wei, Huan Lin, Jialong Tang, Jian Yang, Jianhong Tu, Jianwei Zhang, Jian Yang, Jiaxi Yang, Jingren Zhou, Jingren Zhou, Junyang Lin, Kai Dang, Keqin Bao, Kexin Yang, Le~Yu, Lianghao Deng, Mei Li, Mingfeng Xue, Mingze Li, Pei Zhang, Peng Wang, Qin Zhu, Rui Men, Ruize Gao, Shixuan Liu, Shuang Luo, Tianhao Li, Tianyi Tang, Wenbiao Yin, Xingzhang Ren, Xinyu Wang, Xinyu Zhang, Xuancheng Ren, Yang Fan, Yang Su, Yichang Zhang, Yinger Zhang, Yu~Wan, Yuqiong Liu, Zekun Wang, Zeyu Cui, Zhenru Zhang, Zhipeng Zhou, and Zihan Qiu.
\newblock Qwen3 technical report.
\newblock \emph{CoRR}, abs/2505.09388, 2025.
\newblock \doi{10.48550/ARXIV.2505.09388}.
\newblock URL \url{https://doi.org/10.48550/arXiv.2505.09388}.

\bibitem[Yu et~al.(2025)Yu, Zhang, Zhu, Yuan, Zuo, Yue, Fan, Liu, Liu, Liu, Lin, Lin, Ma, Sheng, Tong, Zhang, Zhang, Zhang, Zhu, Zhu, Chen, Chen, Wang, Yu, Dai, Song, Wei, Zhou, Liu, Ma, Zhang, Yan, Qiao, Wu, and Wang]{DBLP:journals/corr/abs-2503-14476}
Qiying Yu, Zheng Zhang, Ruofei Zhu, Yufeng Yuan, Xiaochen Zuo, Yu~Yue, Tiantian Fan, Gaohong Liu, Lingjun Liu, Xin Liu, Haibin Lin, Zhiqi Lin, Bole Ma, Guangming Sheng, Yuxuan Tong, Chi Zhang, Mofan Zhang, Wang Zhang, Hang Zhu, Jinhua Zhu, Jiaze Chen, Jiangjie Chen, Chengyi Wang, Hongli Yu, Weinan Dai, Yuxuan Song, Xiangpeng Wei, Hao Zhou, Jingjing Liu, Wei{-}Ying Ma, Ya{-}Qin Zhang, Lin Yan, Mu~Qiao, Yonghui Wu, and Mingxuan Wang.
\newblock {DAPO:} an open-source {LLM} reinforcement learning system at scale.
\newblock \emph{CoRR}, abs/2503.14476, 2025.
\newblock \doi{10.48550/ARXIV.2503.14476}.
\newblock URL \url{https://doi.org/10.48550/arXiv.2503.14476}.

\bibitem[Yue et~al.(2025)Yue, Yuan, Yu, Zuo, Zhu, Xu, Chen, Wang, Fan, Du, Wei, Yu, Liu, Liu, Liu, Lin, Lin, Ma, Zhang, Zhang, Zhang, Zhu, Zhang, Liu, Wang, Wu, and Yan]{DBLP:journals/corr/abs-2504-05118}
Yu~Yue, Yufeng Yuan, Qiying Yu, Xiaochen Zuo, Ruofei Zhu, Wenyuan Xu, Jiaze Chen, Cheng{-}Xiang Wang, Tiantian Fan, Zhengyin Du, Xiangpeng Wei, Xiangyu Yu, Gaohong Liu, Juncai Liu, Lingjun Liu, Haibin Lin, Zhiqi Lin, Bole Ma, Chi Zhang, Mofan Zhang, Wang Zhang, Hang Zhu, Ru~Zhang, Xin Liu, Mingxuan Wang, Yonghui Wu, and Lin Yan.
\newblock {VAPO:} efficient and reliable reinforcement learning for advanced reasoning tasks.
\newblock \emph{CoRR}, abs/2504.05118, 2025.
\newblock \doi{10.48550/ARXIV.2504.05118}.
\newblock URL \url{https://doi.org/10.48550/arXiv.2504.05118}.

\end{thebibliography}
\bibliographystyle{iclr2025_conference}

\appendix
\section{Appendix}
\subsection{CISPO Gradient Analysis}
\label{appendix:CISPO}
To better understand the difference between GPPO and the concurrent work CISPO, we analyze the gradient of CISPO in this section.
CISPO builds upon a vanilla REINFORCE objective by incorporating GRPO's relative reward scaling within response groups, while also integrating DAPO's asymmetric clipping strategy on the corrected distribution. The specific loss function takes the following form:
\begin{equation}
\begin{aligned}
\label{cispo_loss_soft}
\mathcal{L}^{\text{CISPO}}(\theta) &= \mathbb{E}_{x\sim\mathcal{D}} \frac{1}{\sum_{j=1}^M T_j} \sum_{j=1}^M \sum_{t=1}^{T_j} \delta\tilde{A}^{(j)} log(\pi_\theta(a_t^{(j)}|s_t^{(j)}).
\end{aligned}
\end{equation}

By expanding the gradient of the CISPO loss function, we obtain the following expression:
\begin{equation}
\begin{aligned}
\label{cispo_gradient_soft}
\nabla_\theta \mathcal{L}^{\text{CISPO}}(\theta) &= \mathbb{E}_{x \sim \mathcal{D}} \left[
\frac{1}{\sum_{j=1}^M T_j} \sum_{j=1}^M \sum_{t=1}^{T_j}
\mathcal{F}_{j,t}(\theta) \cdot \phi_\theta(a_{j,t}, s_{j,t}) \cdot \tilde{A}^{(j)}
\right], \\[1.2em]
\text{where} \quad 
\mathcal{F}_{j,t}(\theta) &= 
\begin{cases}
1-\epsilon_l & \text{if } \delta < 1-\epsilon_l \text{ and } \tilde{A}^{(j)} < 0, \\[0.5em]
1+\epsilon_h & \text{if } \delta > 1+\epsilon_h \text{ and } \tilde{A}^{(j)} > 0, \\[0.5em]
1-\epsilon_l & \text{if } \delta < 1-\epsilon_l \text{ and } \tilde{A}^{(j)} > 0, \\[0.5em]
1+\epsilon_h & \text{if } \delta > 1+\epsilon_h \text{ and } \tilde{A}^{(j)} < 0, \\[0.5em]
\delta & \text{otherwise}.
\end{cases}
\end{aligned}
\end{equation}
As shown in Eq.\ref{grpo_gradient_soft}, GPPO inherits PPO's pessimistic update strategy compared to CISPO. Specifically, when $\delta < 1-\epsilon_l \text{ and } \tilde{A}^{(j)} > 0$, the gradient term $\mathcal{F}_{j,t}(\theta) = 1-\epsilon_l$ in CISPO but remains $\mathcal{F}_{j,t}(\theta) = \delta$ in GPPO, where $0<\delta<\epsilon_l$. This indicates that GPPO distrusts overly optimistic improvements and suppresses such cases through conservative updates, while CISPO applies more aggressive updates with a larger magnitude (i.e., $1-\epsilon_l$). Conversely, when $\delta > 1+\epsilon_l \text{ and } \tilde{A}^{(j)} < 0$, CISPO's gradient term $\mathcal{F}_{j,t}(\theta) = 1+\epsilon_l$, whereas GPPO retains $\mathcal{F}_{j,t}(\theta) = \delta$, where $\delta>\epsilon_h$. Here, GPPO fully trusts negative feedback without suppression, while CISPO applies milder updates (i.e., $1+\epsilon_l$). GPPO's PPO-inherited pessimistic update strategy~\citep{DBLP:journals/corr/SchulmanWDRK17}, ~\textbf{which distrusts excessively optimistic improvements but fully accepts negative feedback, helps prevent policy collapse and enhances training stability.}

\subsection{The General Form of GPPO} 
\label{sec:The General Form of GPPO} 
In this section, we further refine Eq.\ref{grpo_loss_soft} to present a more generalized form of the loss function, which is formulated as follows:
\begin{equation}
\begin{aligned}
\label{grpo_loss_soft_general}
\mathcal{L}^{\text{GPPO}}(\theta) 
&= \mathbb{E}_{x\sim\mathcal{D}} 
   \frac{1}{\sum_{j=1}^M T_j} 
   \sum_{j=1}^M \sum_{t=1}^{T_j} 
   \ell^{(j)} ,
\\[3pt]
\text{where} \quad
\ell^{(j)} &=
\begin{cases}
\beta_1 \cdot \, \dfrac{1-\epsilon_l}{\operatorname{sg}(\delta)} 
         \, \delta \, \cdot \tilde{A}^{(j)}, 
& \text{if } \delta < 1-\epsilon_l \text{ and } \tilde{A}^{(j)} < 0, \\[8pt]
\beta_2 \cdot \, \dfrac{1+\epsilon_h}{\operatorname{sg}(\delta)} 
         \, \delta \, \cdot \tilde{A}^{(j)}, 
& \text{if } \delta > 1+\epsilon_h \text{ and } \tilde{A}^{(j)} > 0 , \\[8pt] 
\delta \cdot \, \tilde{A}^{(j)}, 
& \text{otherwise (i.e., } \delta \tilde{A}^{(j)} \le \text{clip}(\delta, 1-\epsilon_l, 1+\epsilon_h) \cdot \tilde{A}^{(j)}).
\end{cases}
\end{aligned}
\end{equation}
Here, $\beta_1$ and $\beta_2$ are two tunable hyperparameters that control the magnitude of the two clipped boundary losses. In additionl, they influence the scale of gradient backpropagation from the clipped boundaries during optimization. By further expanding the gradient of Eq.\ref{grpo_loss_soft_general}, we obtain the following equation:
\begin{equation}
\begin{aligned}
\label{grpo_gradient_soft_general}
\nabla_\theta \mathcal{L}^{\text{GPPO}}(\theta) &= \mathbb{E}_{x \sim \mathcal{D}} \left[
\frac{1}{\sum_{j=1}^M T_j} \sum_{j=1}^M \sum_{t=1}^{T_j}
\mathcal{F}_{j,t}(\theta) \cdot \phi_\theta(a_{j,t}, s_{j,t}) \cdot \tilde{A}^{(j)}
\right], \\[1.2em]
\text{where} \quad 
\mathcal{F}_{j,t}(\theta) &= 
\begin{cases}
\beta_1 \cdot (1-\epsilon_l) & \text{if } \delta < 1-\epsilon_l \text{ and } \tilde{A}^{(j)} < 0, \\[0.5em]
\beta_2 \cdot (1+\epsilon_h) & \text{if } \delta > 1+\epsilon_h \text{ and } \tilde{A}^{(j)} > 0, \\[0.5em]
\delta & \text{otherwise (i.e., } \delta \tilde{A}^{(j)} \le \text{clip}(\delta, 1-\epsilon_l, 1+\epsilon_h) \cdot \tilde{A}^{(j)}).
\end{cases}
\end{aligned}
\end{equation}
The introduction of $\beta_1$ and $\beta_2$ allows Eq.\ref{grpo_loss_soft_general} to achieve finer-grained gradient control compared to Eq.\ref{grpo_loss_soft}.

\subsection{License for Scientific Artifacts} 
In this paper, we use the VERL\footnote{https://github.com/volcengine/verl} framework for RL training, which was developed by the ByteDance team and is released under the Apache License 2.0.


\end{document}